\documentclass[10pt,twocolumn,letterpaper]{article}

\usepackage[pagenumbers]{cvpr} 

\definecolor{cvprblue}{rgb}{0.21,0.49,0.74}
\usepackage[pagebackref,breaklinks,colorlinks,allcolors=cvprblue]{hyperref}

\usepackage[utf8]{inputenc} 
\usepackage[T1]{fontenc}    
\usepackage{hyperref}       
\usepackage{url}            
\usepackage{booktabs}       
\usepackage{amsfonts}       
\usepackage{nicefrac}       
\usepackage{microtype}      
\usepackage{xcolor}         
\usepackage{graphicx} 
\usepackage{pifont}
\usepackage{adjustbox}
\usepackage{longtable}
\usepackage{svg}
\usepackage{wrapfig}
\usepackage{multirow}
\usepackage{enumitem}
\usepackage{pifont}
\usepackage[table]{xcolor}

\newcommand{\trisign}{\ding{115}}

\newcommand{\limited}{\textcolor{yellow}{\raisebox{-.1em}{\large \trisign}}\kern-0.89em{\scriptsize \textbf{!}} \normalsize}

\newcommand{\limitedl}{\textcolor{yellow}{\raisebox{-.1em}{\large \trisign}}\kern-0.65em{\scriptsize \textbf{!}} \normalsize}

\newcommand{\cmark}{\textcolor{green!60!black}{\ding{51}}}
\newcommand{\xmark}{\textcolor{red}{\ding{55}}}

\usepackage{xspace}
\newcommand{\iqa}{\texttt{VRR-QA}\xspace}

\def\paperID{***} 
\def\confName{CVPR}
\def\confYear{2026}

\title{\iqa: Visual Relational Reasoning in Videos Beyond Explicit Cues}

\author{Sirnam Swetha \quad Rohit Gupta \quad Parth Parag Kulkarni \quad David G Shatwell \quad Jeffrey A Chan Santiago \\ \quad Nyle Siddiqui \quad Joseph Fioresi \quad Mubarak Shah \\
  Institute of Artificial Intelligence, University of Central Florida, USA\\
}

\begin{document}

\twocolumn[{%
\renewcommand\twocolumn[1][]{#1}%
\maketitle
\begin{center}
    \centering
    \captionsetup{type=figure}
    \includegraphics[width=0.96\textwidth]{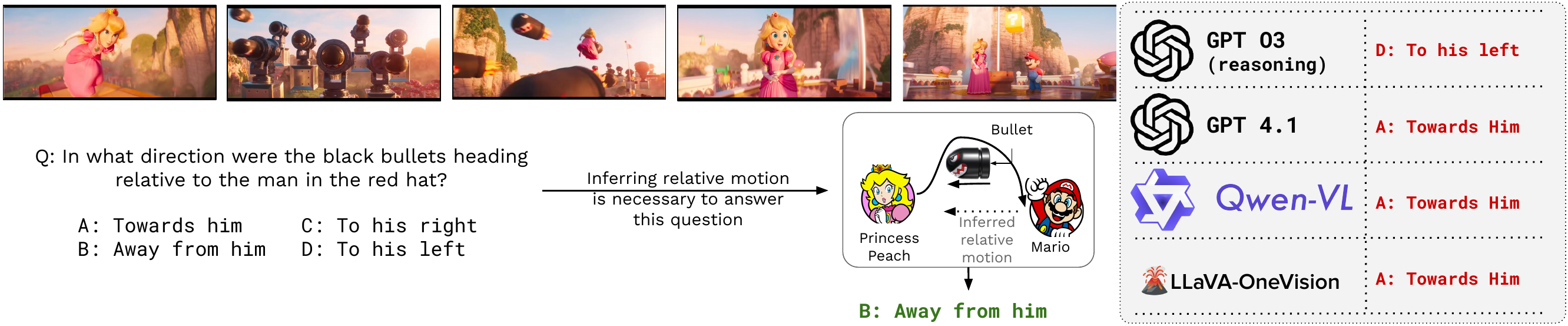}
    \vspace{-0.5em}
    \captionof{figure}{
     Our \underline{V}isual \underline{R}elational \underline{R}easoning benchmark (\iqa) requires implicitly inferring relationships that are not directly visible.  This video shows that black bullets are fired toward Peach and Peach jumps over them to land near Mario. However, it never shows the bullets and Mario together. To answer the question, the bullet's direction relative to Mario must therefore be inferred implicitly based on these cues.
    }
    \label{fig:mario_teaser}
\end{center}
}]

\begin{abstract}
Video Question Answering (VideoQA) has made significant strides by leveraging multimodal learning to align visual and textual modalities. However, current benchmarks overwhelmingly focus on questions answerable through explicit visual content - actions, objects, and events - directly observable within individual frames or short clips. To truly understand videos as humans do, models must go beyond what is directly shown, inferring hidden relationships and contextual cues that are only implied across frames. 
Current benchmarks fail to capture this essential aspect of video understanding.

To address this gap, we introduce \iqa, a benchmark for Visual Relational Reasoning Beyond Explicit Cues. We curate our benchmark from creative and cinematic videos such as movies, that deliberately employ storytelling techniques which omit direct depictions of certain events or relations, requiring viewers to infer them. 
\iqa comprises $1K$ meticulously expert-annotated QA pairs drawn from $1K$ creative video clips covering $15$ genres across $7$ decades of content, from both live-action and animated titles. 
Our extensive evaluations on $14$ leading VideoQA models reveals consistent and significant performance degradation, underscoring their reliance on surface-level visual cues and highlighting the difficulty of implicit reasoning.
Even the best model substantially underperforms human baselines with only 64\% accuracy.
Performance variations across models further illustrate the complexity and diversity of the challenges presented by \iqa. 
By releasing both dataset and data collection framework, \iqa establishes a rigorous, diverse, and reproducible testbed for advancing VideoQA. 
\end{abstract}   
\vspace{-1.5em}
\section{Introduction}
\label{sec:intro}
\vspace{-0.5em}

\begin{figure*}[!t]
    \centering
    \includegraphics[width=0.95\linewidth]{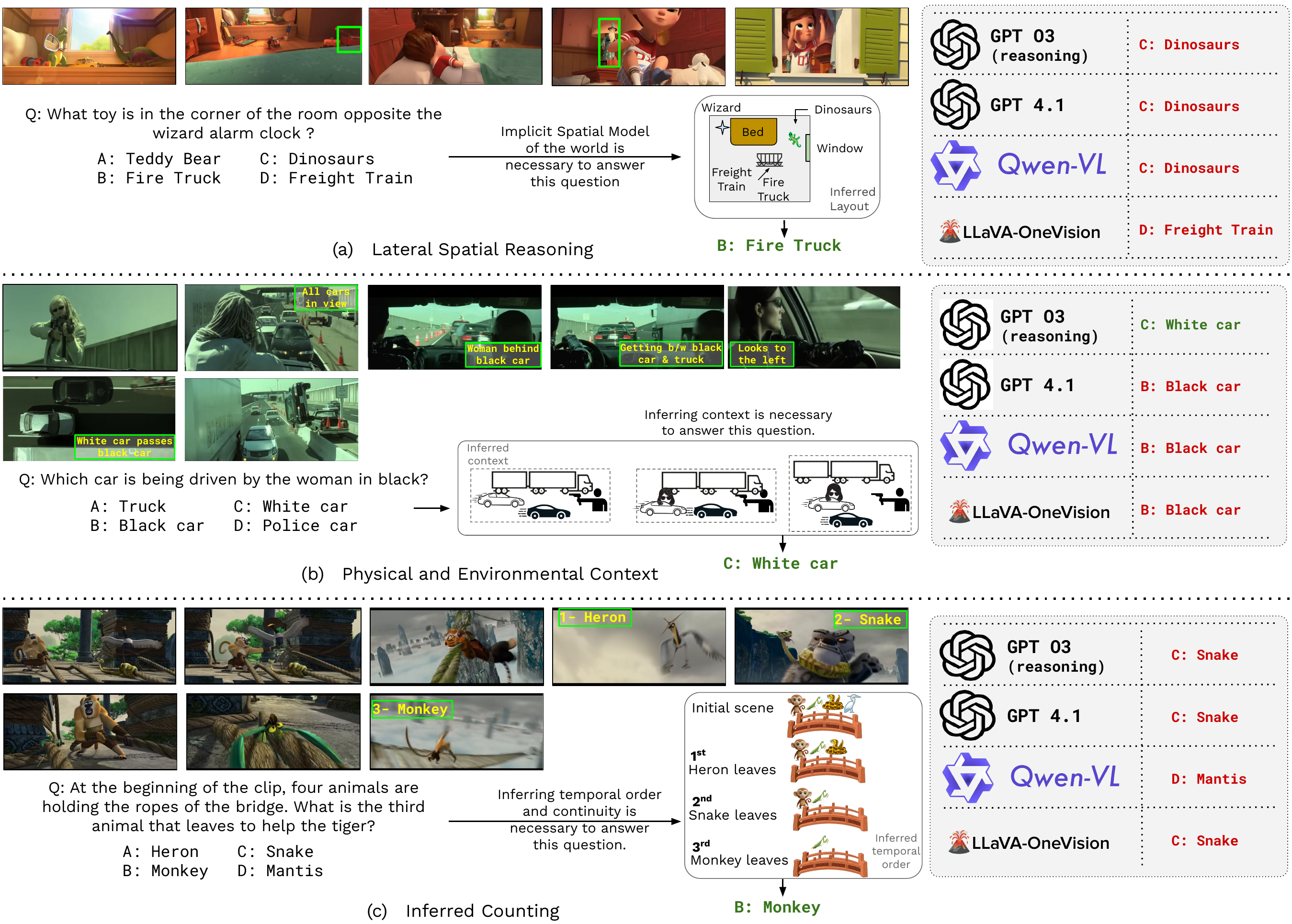}
    \vspace{-1em}
    \caption{\iqa examples, each targeting a distinct implicit‐reasoning dimension. (a) \textbf{Lateral spatial reasoning}-identifying the toy opposite the wizard clock by mentally mapping objects across the scene. (b) \textbf{Physical and Environmental Context}-inferring physical continuity across frames to identify which car is being driven by the woman in black. 
    (c) \textbf{Inferred counting}-determining which animal is the third to leave a bridge by tracking sequential departures that are never fully visible onscreen. Models that excel at explicit perception often fail on these tasks, highlighting the need for benchmarks that probe deeper narrative understanding. }
    \label{fig:teaser}
    \vspace{-1.5em}
\end{figure*}
Video Question Answering (VideoQA) sits at the intersection of computer vision and natural language processing, aiming to answer natural language questions based on video content. Recent progress in VideoQA~\cite{Ren2023TimeChat, zhang2024llavanext-video,Maaz2024VideoGPT+,wang2024internvideo2,li2024llavaOneVision,aria,zhang2024llavavideo, Qwen2-VL,Qwen2.5-VL,yuan2025tarsier2advancinglargevisionlanguage, chen2025scaling, qwen3technicalreport} has been fueled by multimodal learning techniques that integrate visual and textual modalities, enabling impressive performance on datasets where questions are grounded in explicit visual content. These benchmarks~\cite{Xu_2016_CVPR, yu2019activityqa,xiao2021nextqa,lei2018tvqa,wu2021star_situated_reasoning,li2023mvbench, Li2023IntentQA, cai2024temporalbench, fu2024videomme,rawal2024cinepile,liu2024tempcompass,Kesen2024ViLMA, shangguan2024tomato,swetha2025timelogic, Liang2025ReasVQA, Liu2025EntailmentTreeQA, Meng2025OpenO3Video} typically emphasize recognizing objects, identifying actions, and understanding events that are directly observable within individual frames or short clips.

However, human video understanding is not limited to what is explicitly shown. Humans effortlessly infer relationships that unfold implicitly across time: deducing motion direction, building mental spatial maps, tracking continuity across viewpoint changes, or inferring causal links that never appear in the same frame. Figures~\ref{fig:mario_teaser} and \ref{fig:teaser} illustrate this gap: consider the example depicted in Figure~\ref{fig:mario_teaser}, the scene shows the princess Peach running towards Mario (and ending the scene near him), and having to jump over black bullets which are fired towards her, which implies the bullets are moving away from Mario. However, the bullets and Mario are never seen together. The bullet`s direction relative to Mario must therefore be inferred implicitly based these cues. However, existing VideoQA models consistently answer incorrectly, failing to grasp the information across frames. Even the \texttt{GPT-O3} model answers incorrectly, highlighting the widespread difficulty in capturing implicit relationships. 
Leading VideoQA models including GPT-4.1, Qwen-VL, and LLaVA-OneVision consistently fail, often defaulting to answers grounded in local frame content. 

These examples illustrate that current models often struggle with implicit relational reasoning, tending to rely on local frame content.
This challenge becomes even more pronounced in cinematic videos, where implicit reasoning is part of the storytelling norm. As such, movies routinely convey information through off-screen implications, omitted events, viewpoint shifts, or indirect cues. As a result, understanding such content requires integrating cues that are never simultaneously visible, a skill humans perform effortlessly but current models struggle. These properties make cinematic videos an ideal source of naturally occurring implicit relational reasoning tasks. Moreover, such content prevents leakage of visible cues which is common in ego-centric/instructional youtube videos, ensuring that questions truly require reasoning beyond explicit observations.

\begin{table*}[t!]
\centering
\scriptsize
\renewcommand{\arraystretch}{0.9}
\caption{Comparison with existing VideoQA datasets. \iqa uniquely focuses on visual implicit reasoning, annotated fully by authors.}
\begin{tabular}{lcccccccc}
\toprule
\textbf{Dataset} & \multicolumn{2}{c}{\textbf{Tested-Abilities}} & \textbf{Vision} & \textbf{Dynamic} & \textbf{Video Source} & \multicolumn{2}{c}{\textbf{Annotations}} & \textbf{Question} \\
\textbf{} & \textit{Direct} & \textit{Implicit} & \textbf{Only ?} & \textbf{Scene ?} & \textbf{} & \textbf{Humans} & \textbf{Automated} & \textbf{Filtering} \\
\midrule
\textbf{Cinepile}~\cite{rawal2024cinepile} & \cmark & \xmark & \xmark & \cmark & Movie Clips & \xmark & QA (Templates + LLM) & LLM \\
\textbf{TVQA}~\cite{lei2018tvqa} & \cmark & \xmark & \xmark & \cmark & TV Show Clips & QA (Templates)  & - & - \\
\textbf{VideoMME}~\cite{fu2024videomme} & \cmark & \xmark & \xmark & \cmark & Diverse & Experts & - & LLM \\
\textbf{MVBench}~\cite{li2023mvbench} & \cmark & \xmark & \cmark & \cmark & Prior Datasets & Prior Datasets & Wrong Options & - \\
\textbf{TemporalBench}~\cite{cai2024temporalbench} & \cmark & \xmark & \cmark & \cmark & Prior Datasets & Captions & Pairing & LLM \\
\textbf{TempCompass}~\cite{liu2024tempcompass} & \cmark & \xmark & \cmark & \cmark & Stock + Transforms & Class Labels & QA (Templates + LLM) & - \\
\textbf{Open-EQA}~\cite{OpenEQA2023} & \cmark & \xmark  & \cmark & \xmark  & Static 3D Scan & Human & - & 	- \\
\textbf{VSI-Bench}~\cite{yang2025vsibenchthinking} & \cmark & \limited & \cmark & \xmark & Static 3D Scan & QA (Templates) & QA (Templates) & Human\\
\midrule
\textbf{\iqa} & \xmark & \cmark & \cmark & \cmark  & Movie Clips & \multicolumn{3}{c}{\textbf{Experts (Video selection + QA + Filtering)}}  \\
\bottomrule
\end{tabular} 
\label{tab:dataset_comparison}
\vspace{-1.5em}
\end{table*}

To address this critical gap, we introduce \iqa, a benchmark designed to probe the limits of implicit reasoning in VideoQA. We collect $1K$ carefully curated QA pairs from $1K$ diverse video clips sourced from movies. Unlike traditional benchmarks, \iqa focuses exclusively on questions that cannot be answered through direct observation of frames alone. 
Our dataset is organized into $9$ core reasoning categories. We have annotated the data ourselves, thus ensuring that the questions are both challenging and aligned with the nuanced reasoning capabilities we aim to benchmark. Further, we re-verify the annotations amongst ourselves, this curation process also minimizes the ambiguity and guarantees that each question tests meaningful aspects of implicit understanding.

We evaluate prominent VideoQA models on \iqa and observe significant performance drops compared to human-level performance. This finding underscores the current limitations of VideoQA systems, which remain heavily reliant on explicit visual cues. Notably, we find that reasoning-oriented models outperform  non-reasoning models: for example, GPT-o3\cite{openai2024o3o4mini} achieves a 9.8\% higher accuracy than GPT-4.1\cite{openai2024gpt41}. This gap illustrates the necessity of deeper reasoning capabilities to tackle the challenges posed by \iqa, further validating our focus.

In summary, \iqa introduces a new research challenge: to build models capable of implicit reasoning across frames, moving VideoQA closer to true human-like video understanding. Our contributions are listed below:

\begin{itemize}
\item We introduce \iqa, the first benchmark designed to test implicit visual relational  reasoning in VideoQA, focusing on questions that require inference beyond direct visual observations.
\item We manually curate a high-quality dataset of 1k QA pairs across 1k diverse video clips, with annotation conducted by authors to ensure rigor and relevance.
\item We define a taxonomy of 9 categories, covering lateral spatial reasoning, depth and proximity, social dynamics, and more, to facilitate targeted analysis and benchmarking.
\item We benchmark SoTA VideoLLMs on \iqa and reveal a significant gap between current capabilities and true human level understanding.
\end{itemize}
\vspace{-1em}
\section{Related Benchmarks}
\label{sec:relworks}
\vspace{-0.75em}

We compare our proposed dataset against some recent VideoQA benchmarks, which we broadly categorize into Vision-only benchmarks, and Vision and speech benchmarks, which require integration of information from othe modalities like speech.

\begin{figure*}
    \centering
    \includegraphics[width=0.95\linewidth]{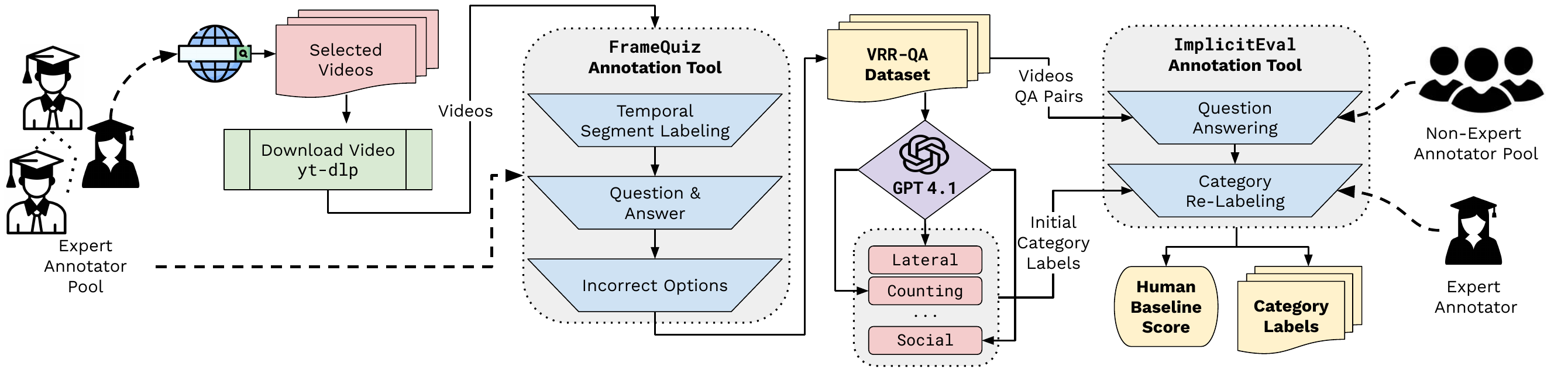}
    \vspace{-1.25em}
    \caption{\iqa Curation Pipeline. We begin by selecting creative video clips and download them. An expert‐annotator pool then uses our FrameQuiz Annotation Tool to (1) mark temporal segments, (2) add a multiple‑choice question and its correct answer for the segment, and (3) craft plausible distractor options. These annotated clips form the raw \iqa Dataset. Next, a non‑expert annotator pool employs the ImplicitEval Annotation Tool to answer each question, yielding a human baseline accuracy score. We run GPT‑4.1\cite{openai2024gpt41} on the dataset to automatically assign initial category tags, which are then relabeled by the expert annotators.}
    \label{fig:pipeline}
    \vspace{-1em}
\end{figure*}

\vspace{-0.25em}
\subsection{Vision Only Benchmarks}
\vspace{-0.25em}

\noindent \textbf{MVBench}~\cite{li2023mvbench} aggregates roughly 4,000 human- or automatically–derived multiple-choice questions drawn from 11 public datasets, pairing each 5–35 s clip with four or five answer options. A ``static-to-dynamic'' pipeline converts image-based tasks into temporally grounded ones, ensuring coverage of both short-term motions and longer-horizon causal phenomena.

\noindent \textbf{TempCompass}~\cite{liu2024tempcompass} focuses on pure temporal manipulation by algorithmically editing 410 royalty-free clips ($\leq$30 s) into pairs/triplets that differ only in one temporal property (e.g., playback reversal, reordered events, speed changes). GPT-3.5 then generates 7,540 diverse tasks (multiple-choice, yes/no, caption matching, constrained captioning). 

\noindent \textbf{TemporalBench}~\cite{cai2024temporalbench} curates 2,032 silent clips (<20 s) from seven public corpora, enriches each with a dense human-written caption plus up to 15 machine-crafted counterfactual captions, and forms 9,867 contrastive pairs targeting fine-grained temporal distinctions (action order, frequency, direction, effector). 

\noindent \textbf{Open-EQA}~\cite{OpenEQA2023} and \textbf{VSIBench}~\cite{yang2025vsibenchthinking} are based on pre-existing 3d scans of indoor environments taken from an egocentric POV. As a result the videos in these datasets carry a dense view of the environment, which needs to be stitched together to answer questions. Due to the dense nature of the sampling, these videos primarily require limited implicit spatial reasoning (indicated by \limitedl   ~symbol). VSI-Bench spans configuration, measurement, and spatiotemporal tasks (e.g., room size, appearance order); OpenEQA covers seven categories (object/attribute/state recognition, localization, spatial and functional reasoning, world knowledge).

\iqa is designed to go beyond existing video benchmarks: Firstly, Implicit, multi-frame inference. Questions demand reasoning about off-screen events, unstated motives, or causal chains spanning multiple clips, no single-frame cues suffice. Secondly, we utilize expert manual curation. Unlike MVBench’s mixed annotations, TempCompass’s algorithmic edits, or TemporalBench’s counterfactual pipeline, our 1K questions from 1k clips are manually authored and rigorously cross-verified by computer-vision experts, ensuring each item truly probes implicit visual reasoning.

\subsection{Vision and Speech Fusion Benchmarks}

\textbf{VideoMME}~\cite{fu2024videomme} comprises 900 expert‐annotated YouTube videos (short <2 min, medium 4–15 min, long 30–60 min) with frames, raw audio, and automatically extracted subtitles. Human annotators author balanced multiple‐choice questions across diverse domains, and a ``text‐only'' filter removes any item solvable without visual or acoustic cues.

\noindent \textbf{TVQA}~\cite{lei2018tvqa} provides $\sim$22K clips (60-90s) from six U.S. TV series, each paired with dialogue subtitles and precise timestamps. Its compositional ``WH‐word ... when/before/after ...'' templates force joint vision–language reasoning and moment localization, but the narrow domain and reliance on spoken dialogue limit purely visual inference.

\noindent \textbf{CinePile}~\cite{rawal2024cinepile} curates $\sim$ 9K movie snippets ($\sim$ 160 s) from MovieClips, augmented with professional audio descriptions and subtitles. An LLM‐driven pipeline yields 304 K multiple‐choice QAs, filtered adversarially for shortcut resistance.

\iqa departs from these benchmarks in three key ways: Firstly, implicit multi‐frame inference: every question demands reasoning about off-screen events, unstated motives, or causal chains not solvable by a single frame or subtitle snippet. Secondly, Visual-only focus: We strip away subtitles and audio tracks entirely, models must extract and integrate visual cues across frames without any textual crutch. Finally, expert manual curation: Unlike datasets that rely on automated pipelines or narrow scripted dialogue, our 1K questions across 1K clips are \textit{manually authored and cross-verified} by computer‐vision experts, ensuring each item truly probes implicit visual reasoning.

\begin{figure*}[t]
    \centering
\begin{subfigure}[]{0.26\textwidth}            
            \includegraphics[width=\textwidth]{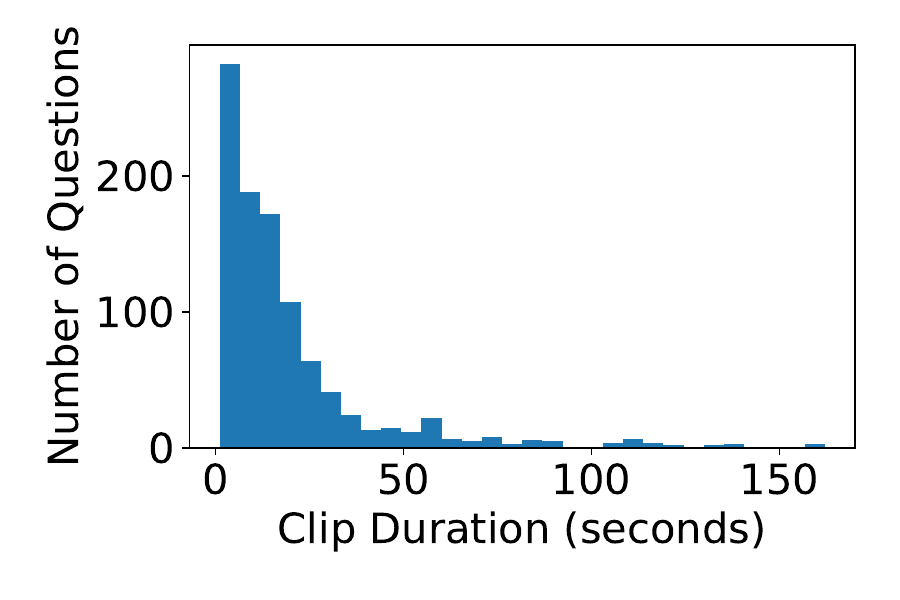}
            \vspace{-2em}
            \caption{Clip Durations}
            \label{fig:clip_dur}
    \end{subfigure}%
    \hfill
    \begin{subfigure}[]{0.26\textwidth}            
            \includegraphics[width=\textwidth]{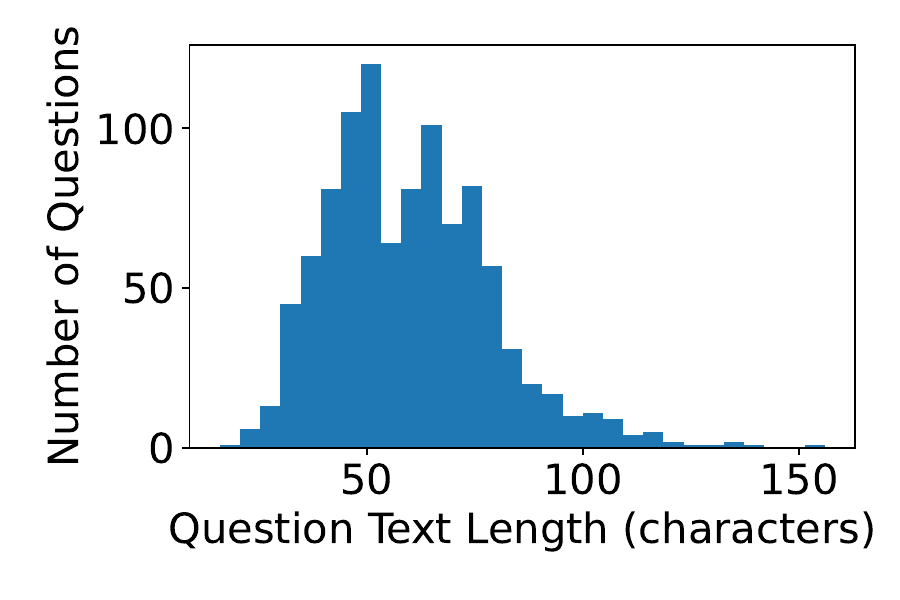}
            \vspace{-2em}
            \caption{Question lengths}
            \label{fig:qlen}
    \end{subfigure}%
    \hfill
    \begin{subfigure}[]{0.26\textwidth}            
            \includegraphics[width=\textwidth]{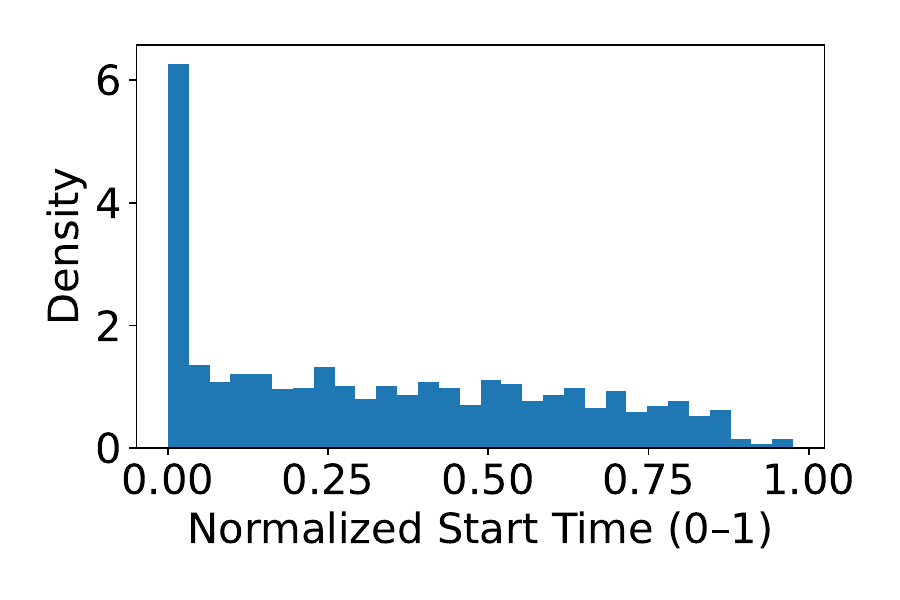}
            \vspace{-2em}
            \caption{Question Start times}
            \label{fig:qstart}
    \end{subfigure}%
    \hfill
    \begin{subfigure}[]{0.19\textwidth}            
            \includegraphics[width=\textwidth]{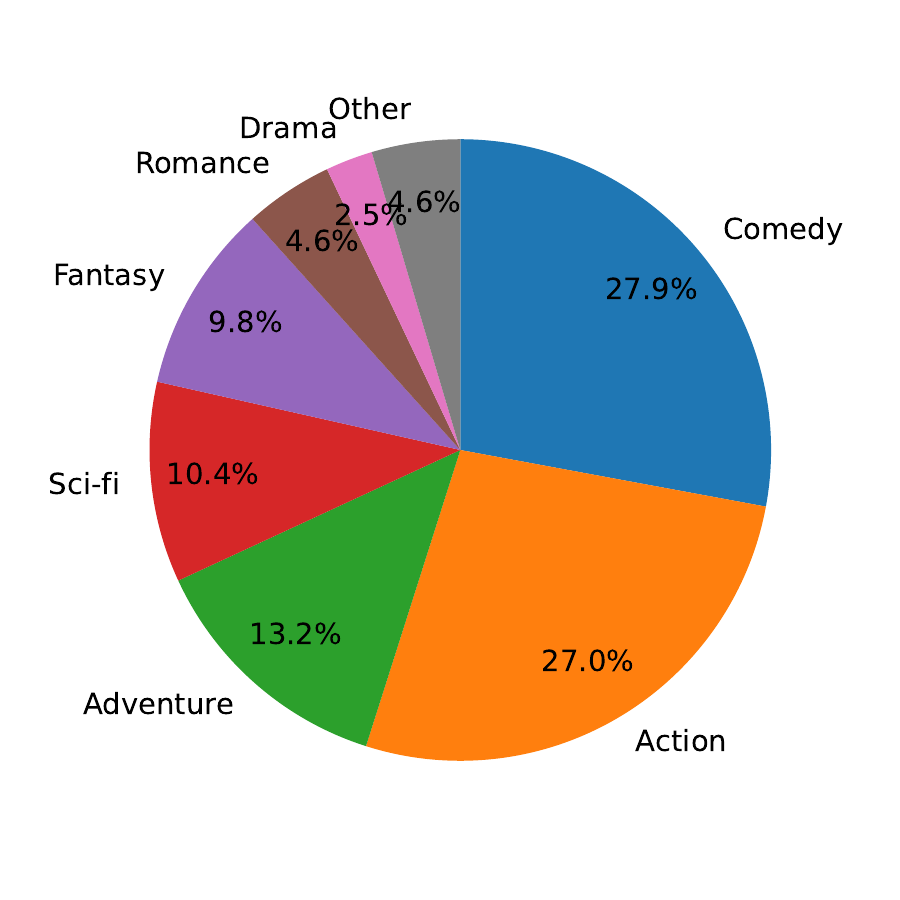}
            \vspace{-2.75em}
            \caption{Distribution of Genres}
            \label{fig:qhist}
    \end{subfigure}%
    
    \vspace{-0.5em}
    \caption{Visualization of \iqa statistics.}
    \label{fig:general_data_stats}
    \vspace{-1em}
\end{figure*}

\section{Dataset Construction}
\label{sec:method}

A critical component of constructing the \iqa benchmark was the creation of high-quality, challenging question-answer pairs that test visual implicit reasoning over video content. To this end, we developed a custom annotation tool specifically designed to streamline and standardize the data collection process for visual implicit reasoning in videos.

\subsection{Annotation Tool and Data Collection}
\label{sec:annotation}

Our custom-built annotation interface was designed to provide an intuitive and efficient workflow. The tool allowed annotators to: 
Watch a video clip directly within the interface, select start and end timestamps marking the temporal window relevant to the annotated question. Write the question and corresponding answer choices, explicitly tying them to the selected video segment. This structured approach ensured that each question was clearly linked to a specific portion of the video, even if the reasoning required drawing upon broader context across multiple scenes. The tool was optimized for fast navigation, enabling annotators to pause, rewind, or step through clips frame-by-frame to closely examine nuanced visual cues. Please refer to Supplementary Section~\ref{sec:supp_ann_tool} for interface figure.

To ensure annotation quality and reliability, we implement a save-and-replay feature within the tool, enabling annotators to revisit their annotated segments, replay the selected video portion, and iteratively refine or validate their annotations before final submission. We also verify the annotations amongst ourselves, please refer to Supplementary Section~\ref{sec:supp_human} for analysis of inter-annotator agreement.  This process ensured that questions accurately aligned with the video context and targeted visual implicit rather than explicit reasoning. We ourselves have annotated questions with the intent to probe deeper inferential reasoning rather than relying solely on directly observable content. Thus the author annotation process contributed to both the conceptual depth and technical relevance of the dataset.

For video selection, we curated a diverse set of 1K creative videos, comprising movies of varied genres and mediums (3D animated and live-action) known for employing narrative techniques such as implied causality, off-screen action, symbolic representation, and indirect storytelling. We prioritized scenes that challenge viewers to make inferences beyond directly visible actions or objects, focusing on content where critical narrative elements are omitted, subtle, or distributed across frames. Statistical characteristics of our collected dataset can be seen in Figure~\ref{fig:general_data_stats}. 

\subsection{Dataset Diversity}

 Benchmarks are meant to evaluate specific skills thoroughly, in a diverse manner. In this section we compare \iqa against related benchmarks on two key aspects of diversity: Question Text Diversity and Video diversity.  We further provide dataset diversity in terms of genre, movie release timeline, media type in Supplementary Section~\ref{sec:supp_detailed_data_stats}.

{
\small
\setlength{\tabcolsep}{1pt}      
\renewcommand{\arraystretch}{0.6} 
\begin{table}[t]
\centering
\begin{tabular}{@{}lc|cc@{}}
\toprule
\textbf{Benchmark} & \textbf{Text MPS} ($\downarrow$) & \# \textbf{Clips} & \textbf{Unique Movies} \\
\midrule
\textcolor{lightgray}{MVBench*}     & \textcolor{lightgray}{0.293}          & \textcolor{lightgray}{3655} & \textcolor{lightgray}{- }  \\
VSI-Bench   & 0.329          & 288  & -   \\
TempCompass$^\dagger$ & 0.228          & 315  & -   \\
Open-EQA    & 0.205          & 180  & -   \\
\midrule
MovieChat   & 0.315          & 170  & $<$100$^\Lambda$ \\
Cinepile    & 0.216          & 148  & 44  \\
MovieQA     & 0.191          & 767  & 26  \\
\textbf{\iqa}        & \textbf{0.161} & 1000  & \textbf{107} \\
\bottomrule
\end{tabular}
\caption{\iqa matches or exceeds existing benchmarks in terms of question text and video diversity. Mean pairwise cosine similarity (MPS) measures question text diversity (lower = more diverse); \textit{Unique} movie sources is a measure of video diversity. *MVBench automatically selects a large \# of clips from existing benchmarks without human curation. $^\dagger$TempCompass has both captioning and QA subtasks, we only include QA for fair comparison. $^\Lambda$ - exact metadata unavailable, as per author communication on \texttt{GitHub}.}
\vspace{-1em}
\label{tab:diversity}
\end{table}
}

\vspace{-1em}
\noindent\paragraph{Question Diversity.} 
To measure question diversity quantitatively, we compute mean pairwise cosine similarity of sentence embeddings between our questions. This metric for text diversity has previously been used in the literature~\cite{tevet-berant-2021-evaluating}. We utilize the \texttt{all-MiniLM-L6-v2} model from Sentence Transformers library to compute text embeddings. 
The results of this analysis are in Table~\ref{tab:diversity}. Lower mean similarity between questions indicates higher diversity, and our questions have higher diversity compared to prior works. The key reason for this is that every question is manually written by expert annotators, without relying on templates or LLMs.

\vspace{-1em}

\noindent\paragraph{Video Diversity.} In recent years, benchmark scales have increased significantly, with bigger benchmarks presumed to be better. However, the primary reason for why scale is important is diversity, and often large VideoQA datasets are constructed from very limited set of videos. Our dataset is built upon 1000 unique clips sourced from 107 movies (see more in Table~\ref{tab:diversity}), which is significantly more diverse than prior movie based QA datasets like Cinepile or MovieQA. These movies span 15 different genres (See Fig.~\ref{fig:general_data_stats}d)

\begin{figure}[h]
    \centering
    \includegraphics[width=0.9\linewidth]{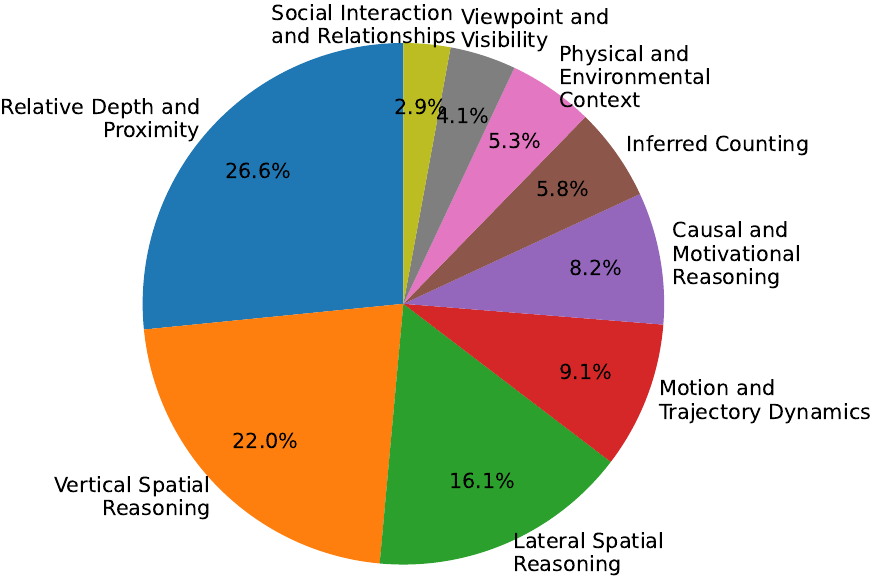}
    \vspace{-0.5em}
    \caption{\iqa distribution across categories}
    \label{fig:pie}
    \vspace{-1em}
\end{figure}

\subsection{Dataset Categorization}

\begin{figure*}[!t]
    \centering
    \includegraphics[width=0.95\linewidth]{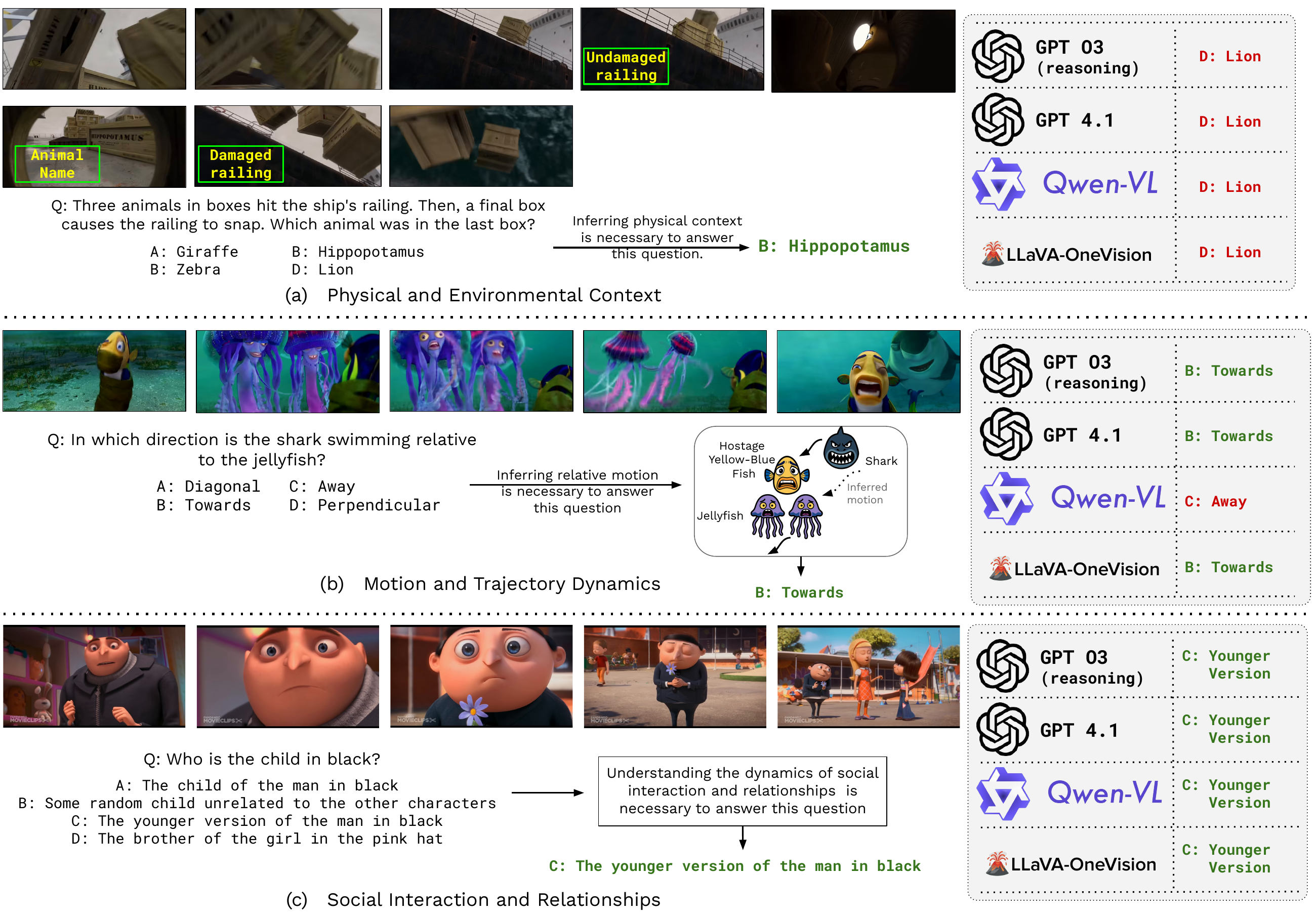}
    \vspace{-0.5em}
    \caption{More Qualitative \iqa examples, targeting distinct implicit‐reasoning dimension. }
    \vspace{-0.5em}
    \label{fig:qual_examples}
\end{figure*}

To ensure comprehensive coverage of the diverse reasoning abilities required for visual relational implicit reasoning, we organize the dataset into nine distinct reasoning categories, each targeting a specific type. Below, we formally define each category of visual implicit reasoning. We present qualitative examples in Figure~\ref{fig:teaser},~\ref{fig:qual_examples} to highlight the implicit nature of questions. We also show differences in temporal length between categories in Supplementary Section~\ref{sec:supp_detailed_data_stats}.

\noindent \textbf{Lateral Spatial Reasoning}
Questions in this category test the ability to infer spatial relationships, positions, or arrangements of objects and characters along lateral orientations. They require viewers to implicitly track or reason about relative positions without explicit directional guidance.

\noindent \textbf{Vertical Spatial Reasoning}
This category assesses the viewer's capacity to implicitly reason about spatial relationships, positions, or arrangements of objects and characters along a vertical axis (above-below orientation). Questions often involve interpreting hierarchical arrangements or vertical positioning that aren't explicitly depicted.

\noindent \textbf{Relative Depth and Proximity}
These evaluate the ability to infer relative distances, and proximity between characters or objects within the scene. They require implicit judgments about which objects or characters are closer or further from the viewer or each other, without explicit depth cues.

\noindent \textbf{Viewpoint and Visibility}
Questions in this category require inferring what is observable from a particular vantage point, whether it be a character’s perspective or camera angle. They must reason about line-of-sight, occlusions, and orientations.

\noindent \textbf{Motion and Trajectory Dynamics}
Questions in this category assess the ability to implicitly track motion, movement directions, and trajectories of characters or objects across discontinuous frames. These movement patterns might be implied and not fully observable in a single scene.

\noindent \textbf{Motivational Reasoning}
Questions in this category require viewers to infer character motives, or likely future events based on incomplete or indirect visual information. 

\noindent \textbf{Inferred Counting}
This category involves implicit counting or enumeration tasks that require aggregating scattered visual evidence across multiple frames or scenes. Such questions demand sustained attention and integration of visual clues over time to infer quantitative details.

\noindent \textbf{Physical and Environmental Context}
Questions in this category probe reasoning about physical elements of the environment, as well as environmental dynamics that may be implied through narrative or partial cues but not shown. 

\noindent \textbf{Social Interaction and Relationships}
This captures reasoning about social dynamics, interactions, and relationships between characters that are inferred through subtle or indirect cues. These questions require understanding of unspoken social behaviors or contextual relational information.

\section{Benchmarking}
\label{sec:exp}

\begin{table*}
\centering
\scriptsize
  \setlength{\tabcolsep}{3pt}
    \renewcommand{\arraystretch}{1.0}
    \caption{Results on \iqa with 16 input frames. \textbf{Best} and \underline{second best} results are highlighted. $^{+}$ - thinking tokens enabled.
    }
\begin{tabular}{l|ccccccccc|c|c} 
\toprule
 Model & \shortstack{Lateral \\ Spatial\\Reasoning} & \shortstack{Vertical \\  Spatial\\Reasoning} & \shortstack{Relative \\ Depth and\\ Proximity} & \shortstack{Viewpoint \\and\\Visibility} & \shortstack{Motion \\\& Traj.\\Dynamics} & \shortstack{Causal \& \\Motivational\\Reasoning} & \shortstack{Inferred\\Counting} & \shortstack{Physical \\ \&  Env.\\Context} & \shortstack{Social  \\ Interaction \& \\Relations} &  Avg. & \shortstack{Macro \\ Avg.} \\ 
\midrule
Human Baseline & 85.4  & 79.1  & 80.4  & 90.0  & 91.9  & 94.4  & 65.9  & 83.3  & 100.0  & 83.0  & 85.6  \\ 
\midrule
\multicolumn{12}{c}{\textbf{Open-Weight Models (7B-Scale)}} \\
\midrule
LLaVA-Next-Video~\cite{zhang2024llavanext-video} & 36.0  & 29.6  & 30.1  & 48.8  & 36.3  & 39.0  & 30.2  & 35.7  & 51.7  & 33.9  & 37.5  \\ 
LLaVA-OneVision~\cite{li2024llavaOneVision}     & 37.3  & 46.8  & 35.0  & 56.1  & 57.1  & 57.3  & 23.3  & 50.0  & 55.2  & \underline{43.4}  & \underline{46.4}  \\ 
LLaVA-Video~\cite{zhang2024llavavideo}   & 36.0  & 44.0  & 31.6  & 56.1  & \textbf{60.4}  & 62.2  & 14.0  & 50.0  & 62.1  & 42.1  & 46.3  \\ 
Gemma 3~\cite{team2025gemma} & \textbf{48.5}  & 38.9  &  32.3  & \textbf{68.3}   & 39.6    &  59.8   &  25.6   &  \textbf{57.1}   & 58.6   &  42.1   & 47.6  \\
InternVL 3~\cite{zhu2025internvl3} & 34.8  & 39.4  & 37.2 &  56.1  &  51.7   & \textbf{64.6}    &   \textbf{34.9}  &  \textbf{57.1}   &  \textbf{75.9}  &  43.3   & \textbf{50.2}  \\ 
Qwen2-VL~\cite{Qwen2-VL} & 39.8  & 46.8  & \textbf{40.6}  & 51.2  & 52.7  & 58.5  & 16.3  & 35.7  & 72.4  & \textbf{44.9}  & 46.0  \\ 
Qwen2.5-VL~\cite{Qwen2.5-VL}  & 41.6  & \textbf{47.2}  & 32.7  & 61.0  & 50.5  & 51.2  & 25.6  & 42.9  & 62.1  & 42.8  & 46.1  \\ 
\midrule
\multicolumn{12}{c}{\textbf{Proprietary Models}} \\
\midrule
Gemini 2.5 Flash~\cite{comanici2025gemini}  & 41.6   & 57.7  & 32.0  & 45 & 60.4  & 71.3  & 42.5  &  69.2 & 81.5 & 49.6  &  55.7 \\
Gemini 3 Flash~\cite{gemini3flash}  & 52.8 & 68.1 & 46.8 & 58.5 & 73.6 & 86.6 & 48.3 & 80.4 & 93.1 & 61.8 & 67.6  \\
Claude 4 Sonnet~\cite{anthropic_claude4_2025_misc}  & 39.1  &  44.9 & 37.9 & 43.9 &  52.8 & 74.4  & 23.3  & 64.3  & 72.4 & 45.4  & 50.3  \\
Claude 4.5 Sonnet~\cite{anthropic_claude4_5_2025_misc} & 42.9 & 42.1 & 36.1 & 46.3 & 45.1 & 62.2 & 15.5 & 60.4 & 65.5 & 42.8 & 46.2 \\
Claude 4.5 Sonnet$^{+}$~\cite{anthropic_claude4_5_2025_misc} & 34.8 & 49.1 & 44.0 & 43.9 & 54.9 & 75.6 & 29.3 & 71.7 & 72.4 & 48.6 & 52.9 \\
GPT 4.1~\cite{openai2024gpt41}& 42.9  & 53.2  & 51.1  & 48.8  & 59.3  & 82.9  & 41.9  & 71.4  & 75.9  & 54.3  & 58.6  \\ 
GPT O3~\cite{openai2024o3o4mini}& 50.3  & 72.2  & 55.3  & 78.0  & 71.4  & 85.4  & 39.5  & 78.6  & 86.2  & 64.1  & 68.6  \\
GPT 5.2~\cite{openai2025gpt52}   & 42.2 & 50.9 & 42.1 & 56.1 & 60.4 & 74.4 & 27.6 & 73.6 & 75.9 & 50.8 & 55.9 \\
GPT 5.2$^{+}$~\cite{openai2025gpt52}  & 38.5 & 64.8 & 45.5 & 53.7 & 68.1 & 84.1 & 39.7 & 79.2 & 86.2 & 56.8 & 62.2 \\
\bottomrule
\end{tabular}

\label{tab:sota}
\end{table*}

\begin{figure*}
    \centering
    \begin{subfigure}[]{0.5\linewidth}
    \includegraphics[width=0.9\linewidth]{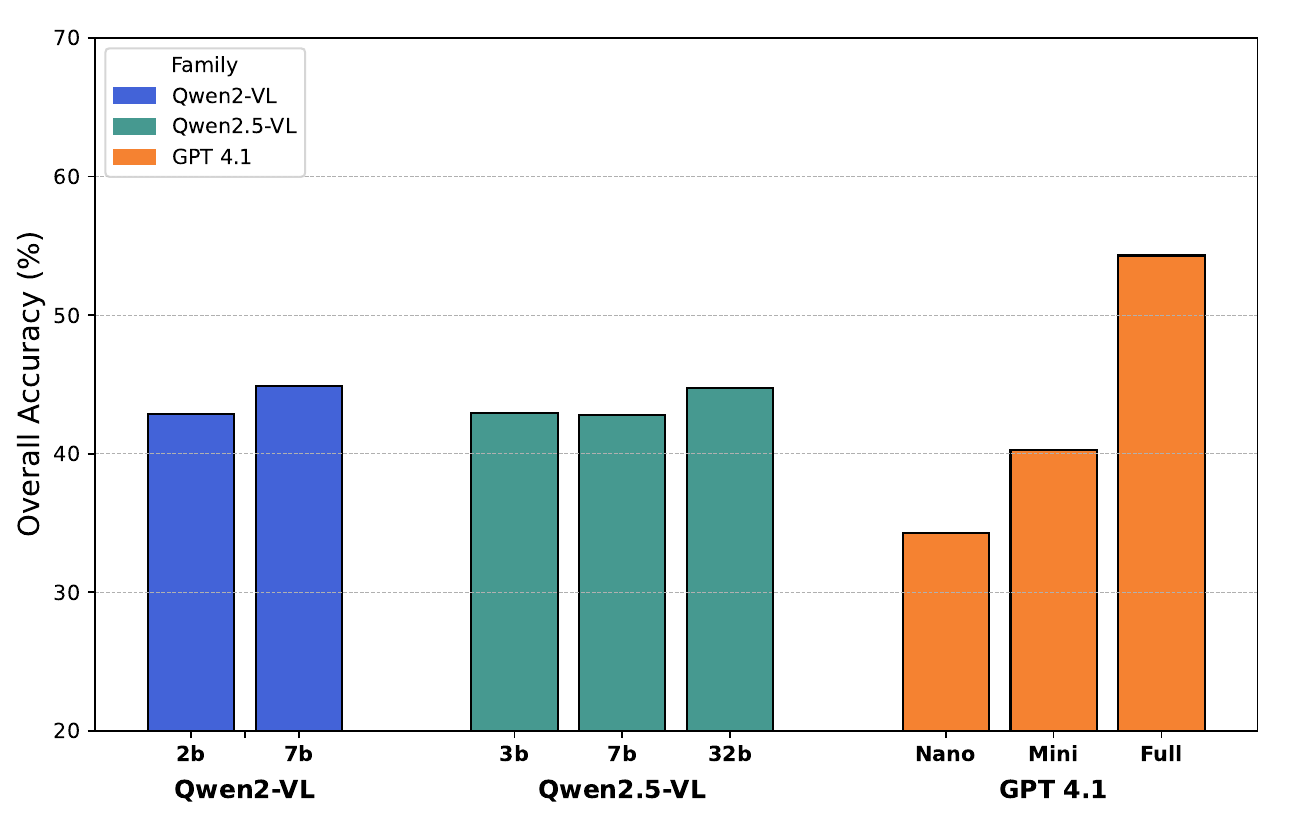}
    \caption{Overall performance vs Model Scale}
    \label{fig:families}
    \end{subfigure}
    \hfill
    \begin{subfigure}[]{0.48\linewidth}
        \includegraphics[width=0.95\linewidth]{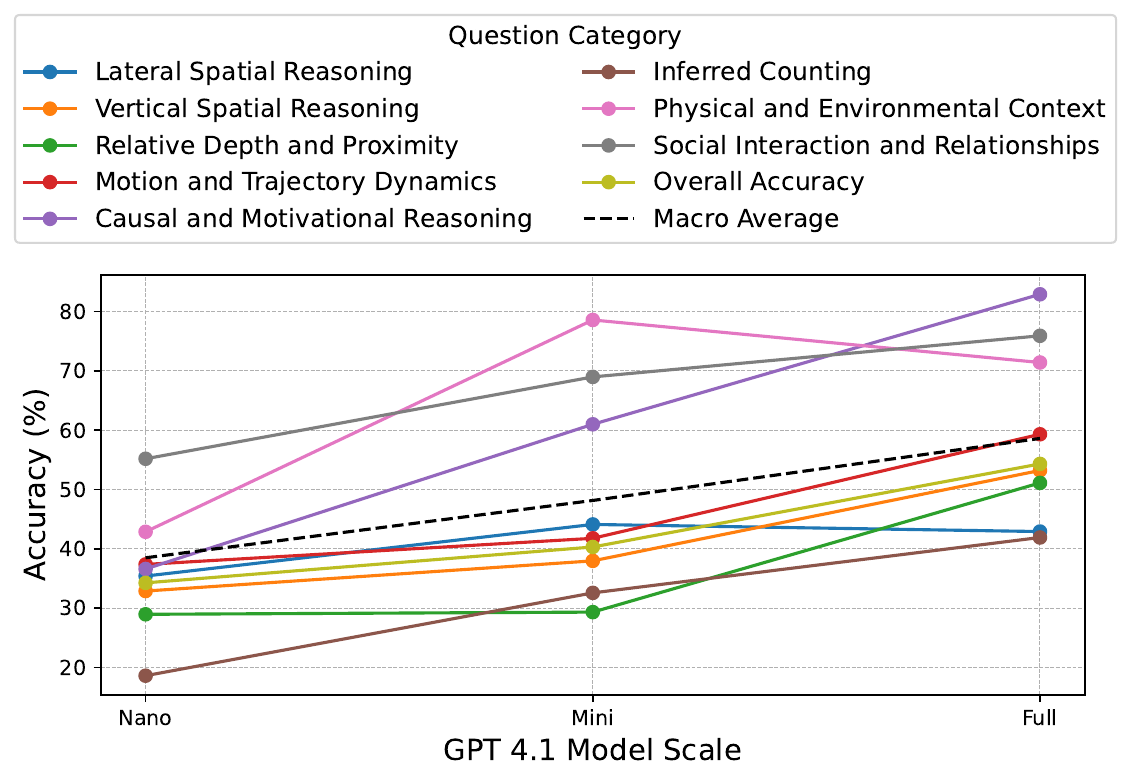}
        \caption{Category-wise performance vs Model Scale}
        \label{fig:det_families}
    \end{subfigure}
    \vspace{-0.5em}
    \caption{Impact of Model Scale.}
    \label{fig:full_families}
    \vspace{-1em}
\end{figure*}

Our evaluation of 30+ VideoQA model configurations across different scales and temporal contexts on the \iqa benchmark reveals several critical insights (see Supplementary Section~\ref{sec:supp_detres} for detailed results).
We evaluate a broad range of open-source and proprietary multimodal models on the \iqa benchmark, focusing on their ability to perform visual implicit reasoning over videos. The evaluation includes multiple model families, scales, and video context lengths.

\noindent We include the following models in our evaluation:
(1) Open-source models: LLaVA-NeXT-Video~\cite{zhang2024llavanext-video}, LLaVA-OneVision~\cite{li2024llavaOneVision}, LLaVA-Video~\cite{zhang2024llavavideo}, Qwen2 VL~\cite{Qwen2-VL}, Qwen2.5 VL~\cite{Qwen2.5-VL}, InternVL3~\cite{zhu2025internvl3}, Gemma 3~\cite{team2025gemma}, 
(2) Closed-source models: GPT-5.2~\cite{openai2025gpt52}, GPT-4.1~\cite{openai2024gpt41}-full, mini, and nano variants, and the reasoning based GPT-O3~\cite{openai2024o3o4mini} model, Gemini 2.5 Flash~\cite{comanici2025gemini}, Gemini 3 Flash~\cite{gemini3flash}, Claude 4 sonnet~\cite{anthropic_claude4_2025_misc}, Claude 4.5 sonnet~\cite{anthropic_claude4_5_2025_misc}. We analyze the performance of these models via category-wise accuracy, and overall average accuracy. We also report the macro-average accuracy, which is the average of category-wise accuracies giving equal weight to each category. Additionally, we explore the impact of integrating an explicit reasoning prompt with GPT models. Please refer to the Supplementary Section~\ref{sec:supp_detres} for detailed scores at model scale and number of frames.

\noindent \textbf{Human Baseline} Non-expert annotators benchmarked themselves against the \iqa dataset using our \texttt{Implicit-Eval} tool. The annotators took around 1 minute per question on average, and achieved an overall score of 83\%, with macro average of 85.6\%.  
The best models lag behind human baseline the most in lateral spatial reasoning, relative depth and motion dynamics.

\paragraph{Overall Performance} Table~\ref{tab:sota} highlights that the reasoning based GPT O3 outperforms all models, achieving 64.1\% overall accuracy and 68.6\% macro average. Among recent models, Gemini 3 Flash performs strongly with 61.8\% overall accuracy, 67.6\% macro average, followed by GPT-5.2$^{+}$ and GPT-4.1, while GPT-5.2 and Gemini 2.5 Flash achieve moderate gains. Claude models generally perform lower, though extended reasoning improves results. Open-source models lag behind proprietary models. The best models, including InternVL-3, Gemma 3, and Qwen variants, achieve moderate performance, while early VideoLLMs such as LLaVA-NeXT-Video perform poorly, highlighting limited capability for implicit reasoning.
Although recent open-source models show similar overall performance, they exhibit clear category-specific strengths; Qwen2-VL, Qwen2.5-VL, Gemma 3, InternVL models excel at spatial reasoning and viewpoint questions, while LLaVA-OneVision based models succeed at Motion, Motivational Reasoning. 
 
Visual implicit reasoning remains an unsolved challenge even for leading VideoQA models. More frames or larger models alone are insufficient, we need architectural innovations or novel training paradigms.
Despite being state-of-the-art in traditional VideoQA, no open-source model crosses 50\% avg. accuracy on \iqa, reflecting the dataset's challenge and the limitations of current architectures in handling implicit, unstated, or cross-frame reasoning.

\noindent \textbf{Model Capacity vs. Visual Implicit Reasoning} GPT models notably benefitted from larger scales, with GPT 4.1~\cite{openai2024gpt41} demonstrating substantial accuracy improvements, indicating more parameters facilitate effective visual implicit reasoning. Among open-source models, larger scales like Qwen2.5-VL~\cite{Qwen2.5-VL}-32b consistently outperformed smaller variants (7b, 3b) by a small margin.

\begin{figure}
    \centering
    \includegraphics[width=\linewidth]{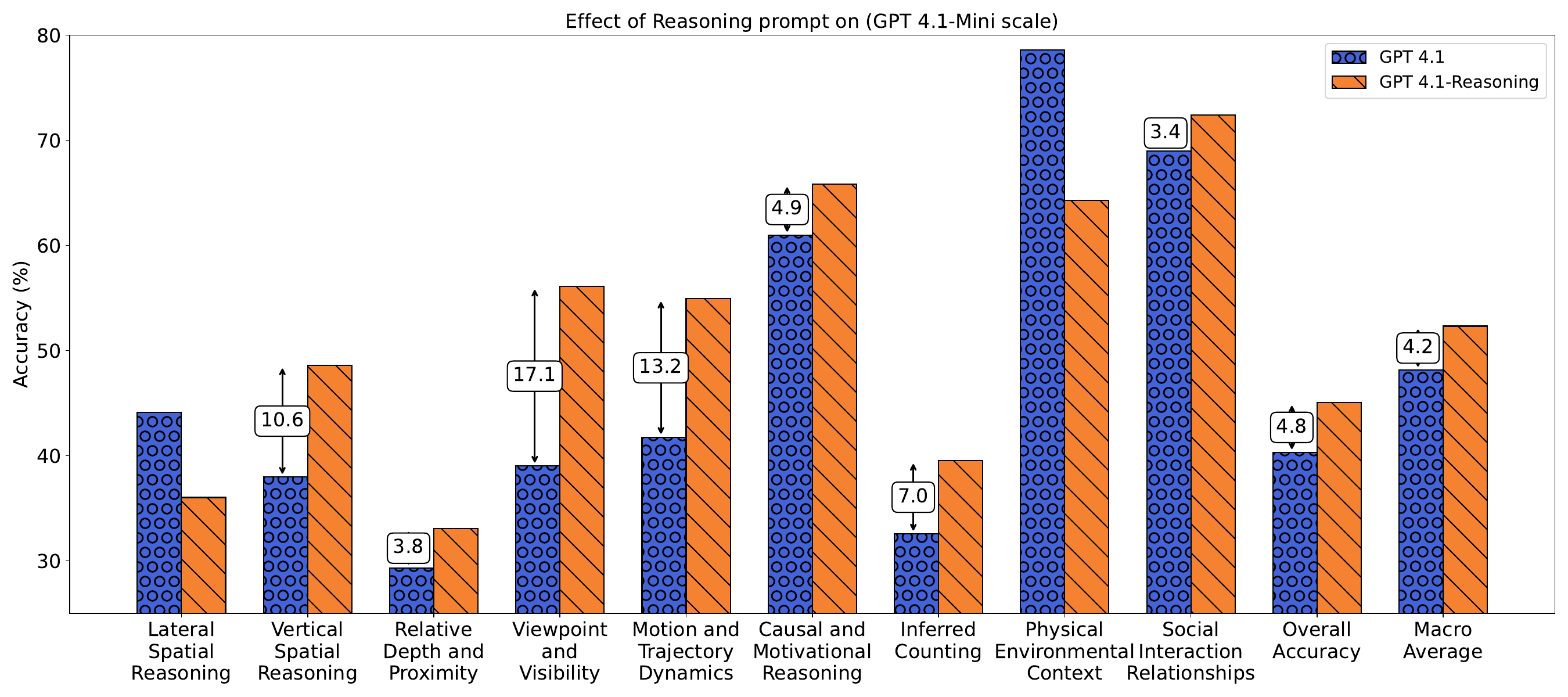}
    \vspace{-0.75em}
    \caption{Effect of Reasoning Prompt}
    \label{fig:reasoning}
    \vspace{-1em}
\end{figure}

\paragraph{Category-wise Analysis}
The performance across distinct implicit reasoning categories reveals varying levels of difficulty as shown in Figure~\ref{fig:full_families} (b),Figure~\ref{fig:reasoning}.\\

\noindent \textbf{(a) Vertical Spatial Reasoning}: Among open-weight models, Qwen2.5-VL performed best with 47.2\% accuracy. GPT O3 excelled with a superior accuracy of 72.2\%.\\   

\noindent \textbf{(c) Motivational Reasoning}: This remains challenging for most models, though proprietary models perform strongly. Gemini 3 Flash achieves the best performance at 86.6\%, closely followed by GPT O3 at 85.4\% and GPT-5.2$^{+}$ at 84.1\%. Open-weight models achieve significantly lower accuracy. \\

\noindent \textbf{(d) Inferred Counting}: This is one of the most challenging categories across all models. While proprietary models perform better, performance remains limited, with Gemini 3 Flash achieving the highest accuracy at 48.3\%, followed by Gemini 2.5 Flash at 42.5\% and GPT-4.1 at 41.9\%. Open-weight models generally perform substantially worse. \\

The experimental results underscore the inherent complexity and nuanced challenges associated with implicit reasoning in videos. They emphasize the clear advantage of proprietary models, such as GPT O3 and Gemini 3 Flash, in handling complex implicit video reasoning tasks, attributed to their larger scale and deeper contextual understanding. Open-source models like Qwen2.5-VL Gemma 3, InternVL 3, while competitive in certain categories, still show substantial room for improvement.

\paragraph{Impact of Reasoning Prompt on GPT Models}
We introduced a structured reasoning prompt specifically emphasizing spatial relationships and narrative summarization before answering questions. GPT 4.1-Mini~\cite{openai2024gpt41}, with the added reasoning prompt, showed enhanced accuracy across multiple categories compared to the standard GPT 4.1-Mini. Specifically, improvements were evident in vertical spatial reasoning (from 37.96\% to 48.61\%), viewpoint and visibility (from 39.02\% to 56.1\%), and motion and trajectory dynamics (from 41.76\% to 54.95\%) as shown in Figure~\ref{fig:reasoning}. This clearly indicates that structured prompting focused on spatial and narrative reasoning significantly enhances model performance. Please refer to Supplementary Section~\ref{sec:supp_reason} for detailed reasoning prompt.\\

\begin{figure}
    \centering
    \includegraphics[width=0.9\linewidth]{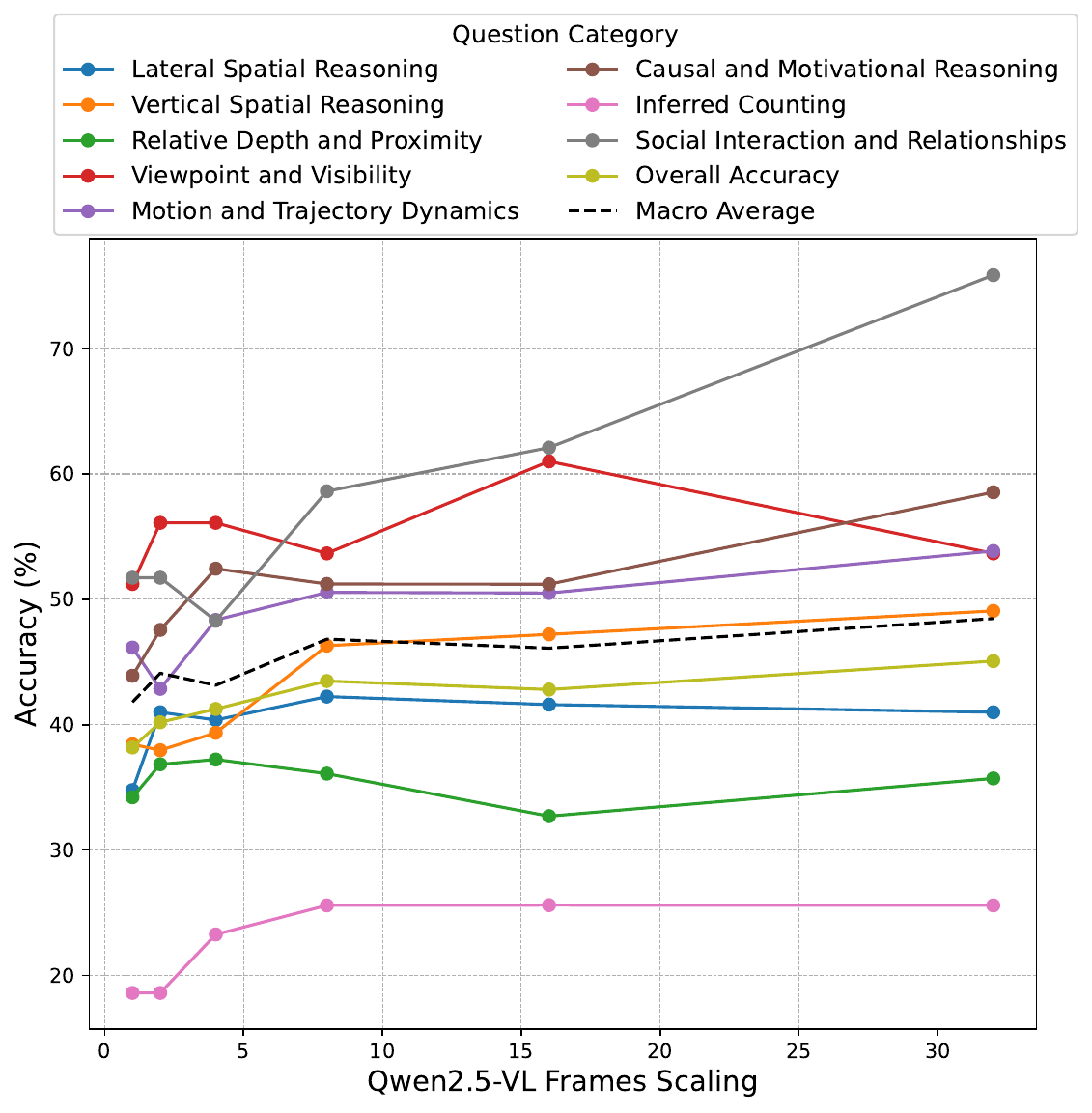}
    \vspace{-0.5em}
    \caption{Effect of Frame Scaling}
    \label{fig:fr_scale}
    \vspace{-1em}
\end{figure}

\noindent \textbf{Effect of Number of Frames:}
Going from 8 to 16 frames improves performance modestly. From 16 frames to 32 frames, accuracy saturates or slightly improves, but not significantly. For Qwen2.5 VL model it improves 2\%. 
\paragraph{Uneven performance across reasoning categories}
Models perform relatively better on Social Interaction and Motivational Reasoning, possibly reflecting pretraining biases towards human-centric scenarios.
Categories involving numerical inference (Inferred Counting) and nuanced spatial reasoning (Relative Depth and Proximity) consistently show lower accuracy, highlighting a need for improvement.

\section{Conclusion}
\label{sec:summ}

We introduce \iqa, a novel benchmark for evaluating visual implicit relational reasoning in videos. 
Across extensive experiments with contemporary models, we find that state-of-the-art systems struggle on visual implicit reasoning: scaling model size and extending temporal context offer only modest gains. Weaknesses are especially pronounced for categories requiring nuanced numerical inference, fine-grained spatial reasoning, and integration of long-term context.
\iqa provides a rigorous evaluation protocol and baseline analyses to catalyze research that moves VideoQA beyond surface-level recognition toward deeper, human-like understanding.

\clearpage

\clearpage
\setcounter{page}{1}
\setcounter{section}{0}
\renewcommand{\thesection}{\Alph{section}}
\maketitlesupplementary

\noindent We organize the supplementary as follows: 

\begin{itemize}
    \item Section~\ref{sec:supp_details}: Data, Tool and Licenses
    \item Section~\ref{sec:supp_detres}: Detailed Results
    \item Section~\ref{sec:supp_detailed_data_stats}: \iqa Detailed Statistics
    \begin{itemize}
        \item \textit{Genres}
        \item \textit{Media Type}
        \item \textit{Movie Release Timeline}
        \item \textit{Difficulty}
        \item \textit{Question Word Distribution}
    \end{itemize}
    \item Section~\ref{sec:supp_reason}: Impact of Reasoning Prompt
    \item Section~\ref{sec:supp_stat}: Experiment Statistical Significance
    \item Section~\ref{sec:supp_human}: Human Baseline
     \begin{itemize}
        \item \textit{Human baselines for Inferred Counting}
    \end{itemize}
    \item Section~\ref{sec:supp_ann_tool}: Annotation Tool Interface
    \item Section~\ref{sec:supp_qual}: Qualitative Results 
    \begin{itemize}
        \item  \textit{Viewpoint and Visibility}
        \item \textit{Physical and Environmental Context}
        \item \textit{Vertical Spatial Reasoning}
        \item \textit{Relative Depth and Proximity}
        \item \textit{Lateral Spatial Reasoning}
        \item \textit{Motion and Trajectory Dynamics}
        \item \textit{Social Interaction and Relationships}
        \item \textit{Inferred Counting}
        \item \textit{Motivational Reasoning}
    \end{itemize}
\end{itemize}

\vspace{-0.5em}
\section{Data, Tool and Licenses}
\label{sec:supp_details}
\vspace{-0.5em}
We publicly release the benchmark, annotation tool, evaluation scripts and evaluation tool with Apache 2.0 license at \href{https://swetha5.github.io/ImplicitQA/}{https://swetha5.github.io/ImplicitQA/}. We ran all our evaluations on NVIDIA A6000 48GB GPUs, and released the eval scripts for reproducibility.

\vspace{-0.5em}
\section{Detailed Results}
\label{sec:supp_detres}
\vspace{-0.5em}

We present detailed results for varying model scales, temporal context across different models in Table~\ref{tab:full_results}. 
The best-performing open-source models were variants of the Qwen2.5 VL\cite{Qwen2-VL} family at 32B scale. Model scale had a noticeable but diminishing effect: Within Qwen2.5 VL, increasing model size from 3B to 32B parameters provided modest performance gains (approximately +2\% accuracy improvement).
Smaller models, such as LLaVA-OneVision, struggled significantly in challenging reasoning categories irrespective of scale. Distinct performance variations across reasoning categories emerged. Social Interaction was relatively easier, with accuracy up to 79.31\% (qwen2.5 VL).

Analysis on increasing frame count generally improves model performance for certain architectures, but this improvement plateaus or slightly degrades beyond a threshold. For instance, LLaVA-NeXT-Video\cite{zhang2024llavanext-video} exhibited a peak performance at 16 frames (33.9\%) with a marginal decrease at 32 frames (32.56\%). Similarly, LLaVA-OneVision\cite{li2024llavaOneVision} performed slightly better at 32 frames (43.16\%) compared to 16 frames (43.4\%), indicating negligible performance gains. For Qwen2.5-VL\cite{Qwen2.5-VL} 32B model, increasing frames from 4 to 32 resulted in accuracy improvements, particularly evident in the inferred counting \& social interaction categories, achieving substantial accuracy improvements from 27.91\% to 41.86\% and from 62.07\% to 79.31\%, respectively. This suggests deeper frame context substantially aids in specific visual implicit reasoning tasks.

\begin{table*}[htbp]
\centering
\caption{Detailed Results on \iqa for all visual relational implicit reasoning categories on various VideoLMMs in multiple settings.}
\vspace{1em}
\resizebox{\textwidth}{!}{
\begin{tabular}{l|c|c|c|c|c|c|c|c|c|c|c|c|c}
\toprule

Model  & Scale & \#Frames & \shortstack{Lateral \\ Spatial\\Reasoning} & \shortstack{Vertical \\  Spatial\\Reasoning} & \shortstack{Relative \\ Depth and\\ Proximity} & \shortstack{Viewpoint \\and\\Visibility} & \shortstack{Motion \\\& Traj.\\Dynamics} & \shortstack{\\ Motivational\\Reasoning} & \shortstack{Inferred\\Counting} & \shortstack{Physical \\ \&  Env.\\Context} & \shortstack{Social  \\ Interaction \& \\Relations} &  Avg. & \shortstack{Macro \\ Avg.} \\ 
\midrule
\multirow{3}{*}{LLaVA-NeXT-Video~\cite{zhang2024llavanext-video}} & 7b & 8 & 34.78 & 31.48 & 28.95 & 43.90 & 37.36 & 40.24 & 23.26 & 42.86 & 55.17 & 33.72 & 37.56 \\
 & 7b & 16 & 36.00 & 29.60 & 30.10 & 48.80 & 36.30 & 39.00 & 30.20 & 35.70 & 51.70 & 33.90 & 37.50 \\
 & 7b & 32 & 34.78 & 29.63 & 30.08 & 39.02 & 36.26 & 36.59 & 27.91 & 35.71 & 37.93 & 32.56 & 34.21 \\
\midrule
\multirow{2}{*}{LLaVA-OneVision~\cite{li2024llavaOneVision}} & 7b & 16 & 37.30 & 46.80 & 35.00 & 56.10 & 57.10 & 57.30 & 23.30 & 50.00 & 55.20 & 43.40 & 46.40 \\
 & 7b & 32 & 35.40 & 50.46 & 33.08 & 53.66 & 56.04 & 56.10 & 18.60 & 50.00 & 65.52 & 43.16 & 46.54 \\
\midrule
\multirow{3}{*}{LLaVA-Video~\cite{zhang2024llavavideo}} & 7b & 8 & 31.68 & 41.20 & 29.32 & 48.78 & 57.14 & 57.32 & 13.95 & 50.00 & 62.07 & 39.02 & 43.50 \\
 & 7b & 16 & 36.00 & 44.00 & 31.60 & 56.10 & 60.40 & 62.20 & 14.00 & 50.00 & 62.10 & 42.10 & 46.30 \\
 & 7b & 32 & 36.02 & 47.22 & 32.71 & 51.22 & 58.24 & 63.41 & 16.28 & 57.14 & 68.97 & 43.27 & 47.91 \\
\midrule

\multirow{11}{*}{Qwen2.5-VL~\cite{Qwen2.5-VL}}  & 3b & 16 & 39.75 & 43.98 & 33.83 & 63.41 & 52.75 & 56.10 & 20.93 & 57.14 & 65.52 & 42.95 & 48.16 \\
& 7b & 1 & 34.78 & 38.43 & 34.21 & 51.22 & 46.15 & 43.90 & 18.60 & 57.14 & 51.72 & 38.18 & 41.80 \\
& 7b & 2 & 40.99 & 37.96 & 36.84 & 56.10 & 42.86 & 47.56 & 18.60 & 64.29 & 51.72 & 40.19 & 44.10 \\
& 7b & 4 & 40.37 & 39.35 & 37.22 & 56.10 & 48.35 & 52.44 & 23.26 & 42.86 & 48.28 & 41.25 & 43.14 \\
& 7b & 8 & 42.24 & 46.30 & 36.09 & 53.66 & 50.55 & 51.22 & 25.58 & 57.14 & 58.62 & 43.48 & 46.82 \\
& 7b & 16 & 41.60 & 47.20 & 32.70 & 61.00 & 50.50 & 51.20 & 25.60 & 42.90 & 62.10 & 42.80 & 46.10 \\
& 7b & 32 & 40.99 & 49.07 & 35.71 & 53.66 & 53.85 & 58.54 & 25.58 & 42.86 & 75.86 & \textbf{45.07} & 48.46 \\
& 32b & 4 & 39.75 & 44.91 & 39.10 & 48.78 & 41.76 & 57.32 & 27.91 & 57.14 & 62.07 & 43.27 & 46.53 \\
& 32b & 8 & 38.51 & 45.83 & 40.98 & 56.10 & 47.25 & 59.76 & 23.26 & 50.00 & 62.07 & 44.54 & 47.08 \\
& 32b & 16 & 38.51 & 48.15 & 37.59 & 51.22 & 49.45 & 62.20 & 32.56 & 50.00 & 62.07 & 44.75 & 47.97 \\
& 32b & 32 & 39.75 & 45.83 & 35.71 & 43.90 & 49.45 & 64.63 & 41.86 & 57.14 & 79.31 & 44.86 & \textbf{50.84} \\
\midrule
\multirow{7}{*}{Qwen2-VL~\cite{Qwen2-VL}} & 2b & 16 & 34.78 & 43.98 & 47.37 & 60.98 & 39.56 & 43.90 & 16.28 & 57.14 & 51.72 & 42.84 & 43.97 \\
 & 7b & 1 & 38.51 & 39.35 & 36.84 & 53.66 & 42.86 & 48.78 & 13.95 & 42.86 & 55.17 & 39.66 & 41.33 \\
 & 7b & 2 & 36.65 & 40.28 & 40.60 & 46.34 & 50.55 & 50.00 & 20.93 & 42.86 & 58.62 & 41.57 & 42.98 \\
 & 7b & 4 & 39.13 & 45.83 & 39.47 & 46.34 & 49.45 & 51.22 & 20.93 & 50.00 & 58.62 & 43.05 & 44.56 \\
 & 7b & 8 & 37.27 & 47.69 & 40.98 & 53.66 & 48.35 & 54.88 & 16.28 & 50.00 & 65.52 & 44.11 & 46.07 \\
 & 7b & 16 & 39.80 & 46.80 & 40.60 & 51.20 & 52.70 & 58.50 & 16.30 & 35.70 & 72.40 & 44.90 & 46.00 \\
 & 7b & 32 & 40.37 & 49.07 & 40.98 & 48.78 & 48.35 & 60.98 & 16.28 & 35.71 & 62.07 & 44.96 & 44.73 \\
\midrule
\multirow{5}{*}{LongVILA-R1~\cite{chen2025scaling}} & 7b & 8 & 39.13 & 24.09 & 26.69 & 48.78 &  42.86& 48.78 & 22.41 & 49.06 & 41.38 & 33.67 & 38.13 \\
& 7b & 16 & 35.40 & 28.18 & 30.08 & 39.02 & 42.86 & 52.44 & 31.03 & 50.94 & 58.62 & 35.86 & 40.95 \\
& 7b & 32 & 34.78 & 25.00 & 29.32 & 41.46 & 42.86 & 56.10 & 31.03 & 52.83 & 51.72 & 35.16  & 40.57 \\
& 7b & 64 & 34.16 & 28.64 & 26.69 & 53.66 & 45.05 & 46.34 & 29.31 & 49.06 & 48.28 & 34.67  & 40.13 \\
& 7b & 128 & 26.71 & 26.82 & 25.19 & 46.34 & 43.96 & 48.78 & 27.59 & 49.06 & 51.72 & 32.47 & 38.46 \\
\multicolumn{14}{c}{\textbf{Proprietary Models}} \\
\midrule
\multirow{3}{*}{GPT 4.1~\cite{openai2024gpt41}} & Full & 16 & 42.90 & 53.20 & 51.10 & 48.80 & 59.30 & 82.90 & 41.90 & 71.40 & 75.90 & 54.30 & 58.60 \\
 & Mini & 16 & 44.10 & 37.96 & 29.32 & 39.02 & 41.76 & 60.98 & 32.56 & 78.57 & 68.97 & 40.30 & 48.14 \\
 & Nano & 16 & 35.40 & 32.87 & 28.95 & 58.54 & 37.36 & 36.59 & 18.60 & 42.86 & 55.17 & 34.25 & 38.48 \\
\midrule
GPT O3~\cite{openai2024o3o4mini} & Full & 16 & 50.30 & 72.20 & 55.30 & 78.00 & 71.40 & 85.40 & 39.50 & 78.60 & 86.20 & \textbf{64.10} & \textbf{68.60} \\
\bottomrule
\end{tabular}
}

\label{tab:full_results}
\end{table*}

\begin{figure*}[t]
    \centering
    \begin{subfigure}[]{0.4\textwidth}            
            \includegraphics[width=\textwidth]{images/distribution_pGenre.pdf}
            \vspace{-2em}
            \caption{Primary Genre Distribution}
            \label{fig:genre}
    \end{subfigure}%
    \begin{subfigure}[]{0.4\textwidth}            
            \includegraphics[width=\textwidth]{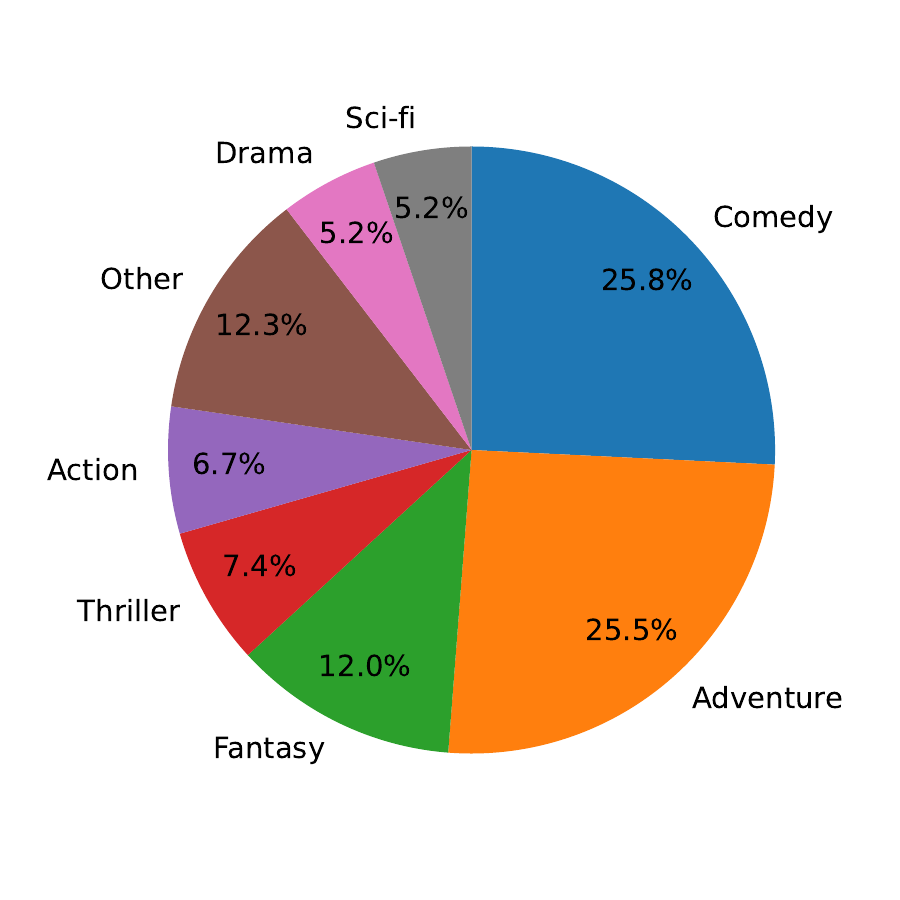}
            \vspace{-2em}
            \caption{Secondary Genre Distribution}
            \label{fig:sgenre}
    \end{subfigure}%
    \hfill
    \caption{Visualization of \iqa video statistics across primary and secondary genres for the top seven most frequent categories. A large proportion of animation videos fall under the Comedy genre, contributing to the higher number of samples annotated as Comedy in both primary and secondary genre distributions.}
    \label{fig:supp_genre_data_stats}
\end{figure*}

\vspace{-0.5em}
\section{\iqa Detailed Statistics}
\label{sec:supp_detailed_data_stats}
\vspace{-0.5em}
This section provides additional statistics for \iqa benchmark, highlighting the dataset's diversity across multiple dimensions, including 
\begin{itemize}
    \item Genre
    \item Media type
    \item Movie Release Timeline
    \item Difficulty based on hard-ness score
\end{itemize}

\vspace{-0.5em}
\subsection{Genres} 

\vspace{-0.5em}
To further characterize the content diversity in \iqa, we manually annotate both the primary and secondary genres of each video. We assign these genre annotations by considering genres listed on dedicated pages for each movie on publicly available sources such as IMDb\footnote{https://www.imdb.com/} and Wikipedia\footnote{https://www.wikipedia.org/}. 

We assign each video a primary genre, which represents a movie's core theme and structure; and a secondary genre which reflects additional aspects of a movie. We observe that these primary and secondary genres come from a total of 15 different genres which are listed(in alphabetical order):
\begin{itemize}
    \item Action 
    \item Adventure
    \item Black comedy
    \item Comedy
    \item Crime 
    \item Drama
    \item Fantasy 
    \item Horror
    \item Mystery 
    \item Psychological horror/thriller
    \item Romance 
    \item Sci-fi
    \item Socio-political 
    \item Thriller
    \item Western
\end{itemize}

\noindent Figure~\ref{fig:supp_genre_data_stats} shows the primary and secondary genre distribution of our dataset. We observe that the dataset includes a wide range of genres, with Comedy, Action, and Adventure being the most prominent primary,  while Comedy, Adventure and Fantasy being the top 3 secondary. This broad genre coverage ensures the benchmark captures diverse narrative structures, thematic elements, and stylistic conventions - essential for evaluating visual implicit reasoning across contexts.

\begin{figure}
    \centering
    \includegraphics[width=\linewidth]{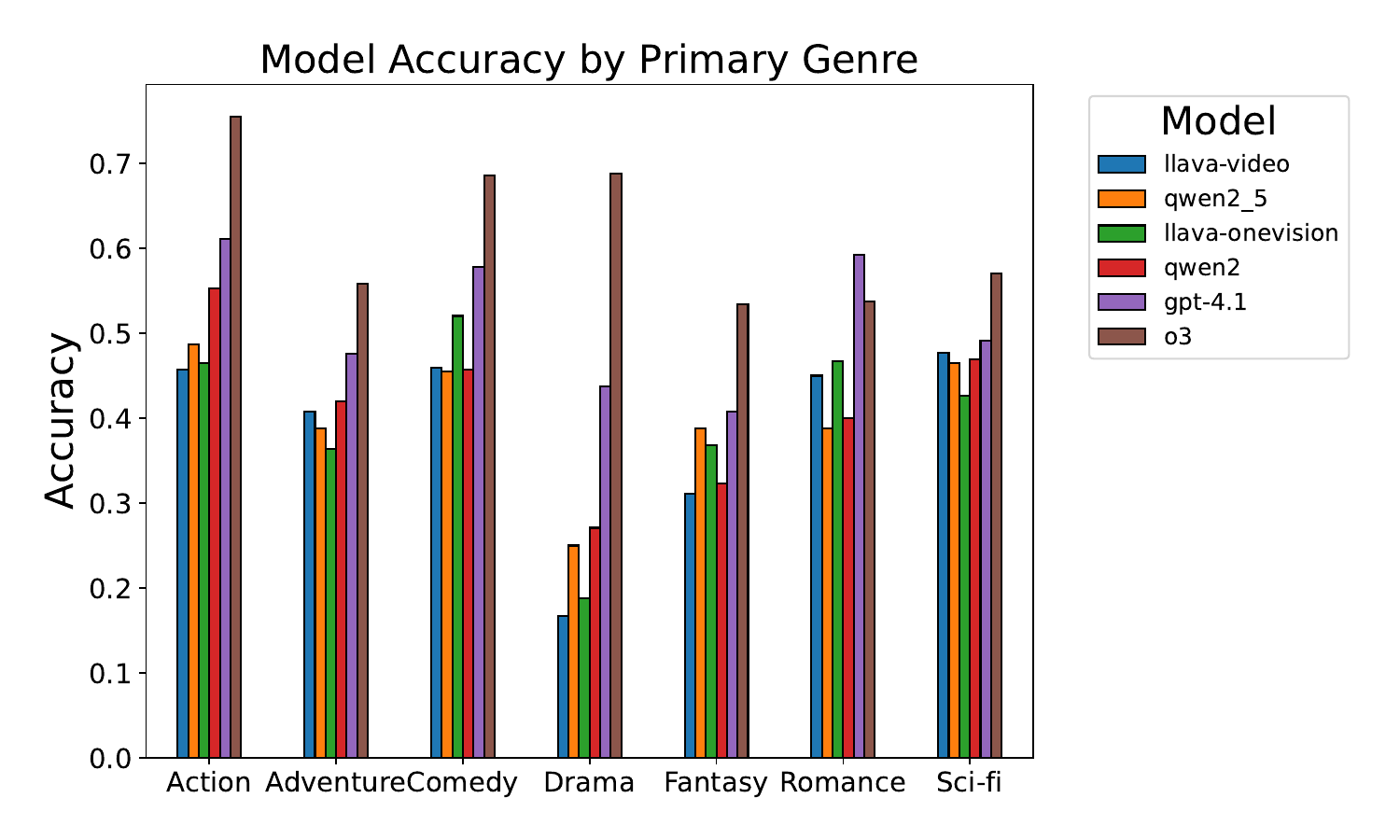}
    \caption{Model accuracy across primary video genres in the \iqa\ dataset. Performance varies significantly by genre, with O3 model consistently leading across genres except Romance.}
    \label{fig:supp_genreresult}
\end{figure}

To investigate how genre influences model performance, we show accuracy across primary genre categories in \iqa. As shown in Figure~\ref{fig:supp_genreresult}, genre plays a substantial role in performance variation. Overall models perform best on Action, Comedy, and Romance. In contrast, performance drops for genres like Drama and Fantasy. Notably, the O3 model outperforms all others across every genre except Romance, suggesting its stronger ability to generalize across narrative structures. The variation across genres also underscores the importance of content diversity in benchmark design, as genre-specific reasoning challenges reveal gaps in current video LMMs capabilities.

\begin{figure}
    \centering
    \includegraphics[width=\linewidth]{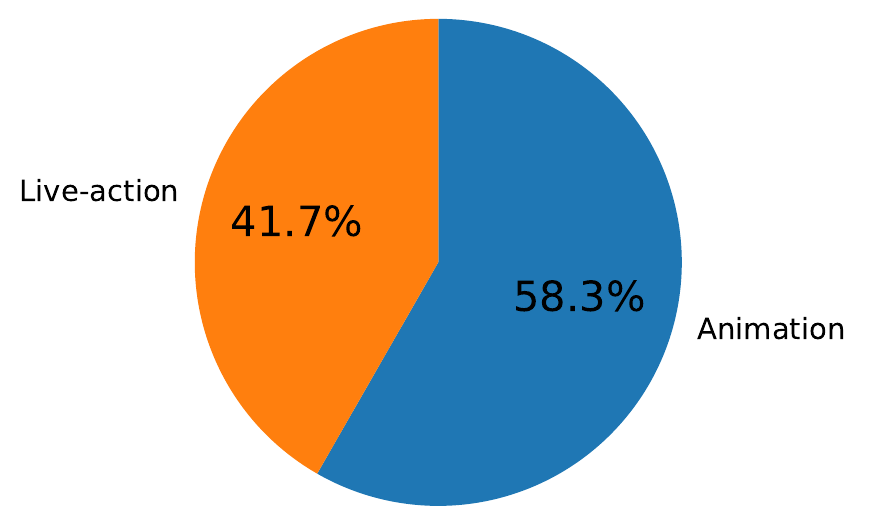}
    \caption{Distribution of Media Type in \iqa.}
    \label{fig:supp_media_stats}
\end{figure}
\subsection{Media Type}
We further categorize the videos into Live-Action and Animation to highlight the diversity in visual domains present in \iqa. As shown in Figure~\ref{fig:supp_media_stats}, the dataset maintains a balanced composition across both categories, with 58.3\% of the videos being animated and 41.7\% live-action. This mix ensures exposure to varied stylistic, motion, and rendering characteristics that challenges LMMs.

\begin{figure}
    \centering
    \includegraphics[width=0.9\linewidth]{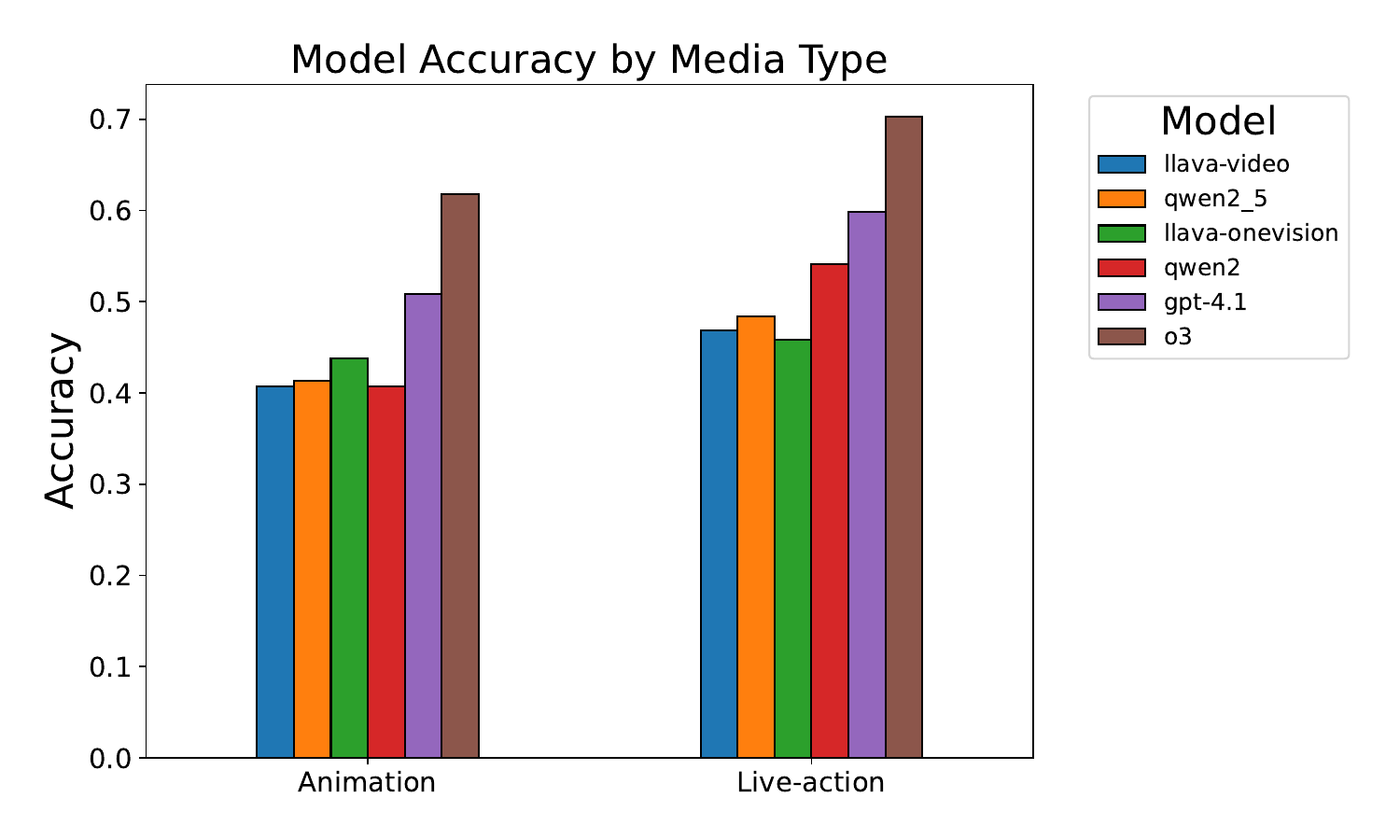}
    \caption{Model accuracy across media types (Animation vs. Live-action). Performance is consistently higher on live-action videos, with the largest gains observed in higher-capacity models such as GPT-4.1 and O3. }
    \label{fig:supp_animationresult}
\end{figure}

To further understand model generalization across different visual domains, we show performance on animation and live-action videos. As shown in Figure~\ref{fig:supp_animationresult}, all models demonstrate stronger performance on live-action content. The gap is especially more for larger models like GPT-4.1 and O3, which outperform others by a substantial margin. These results indicate that models may rely more effectively on grounded visual signals and realistic spatial cues present in live-action videos, whereas stylized representations in animation pose additional challenges for visual relational implicit reasoning. This highlights the need for further adaptation for animation-rich inputs.

\begin{figure}[t]
    \centering
    \includegraphics[width=\linewidth]{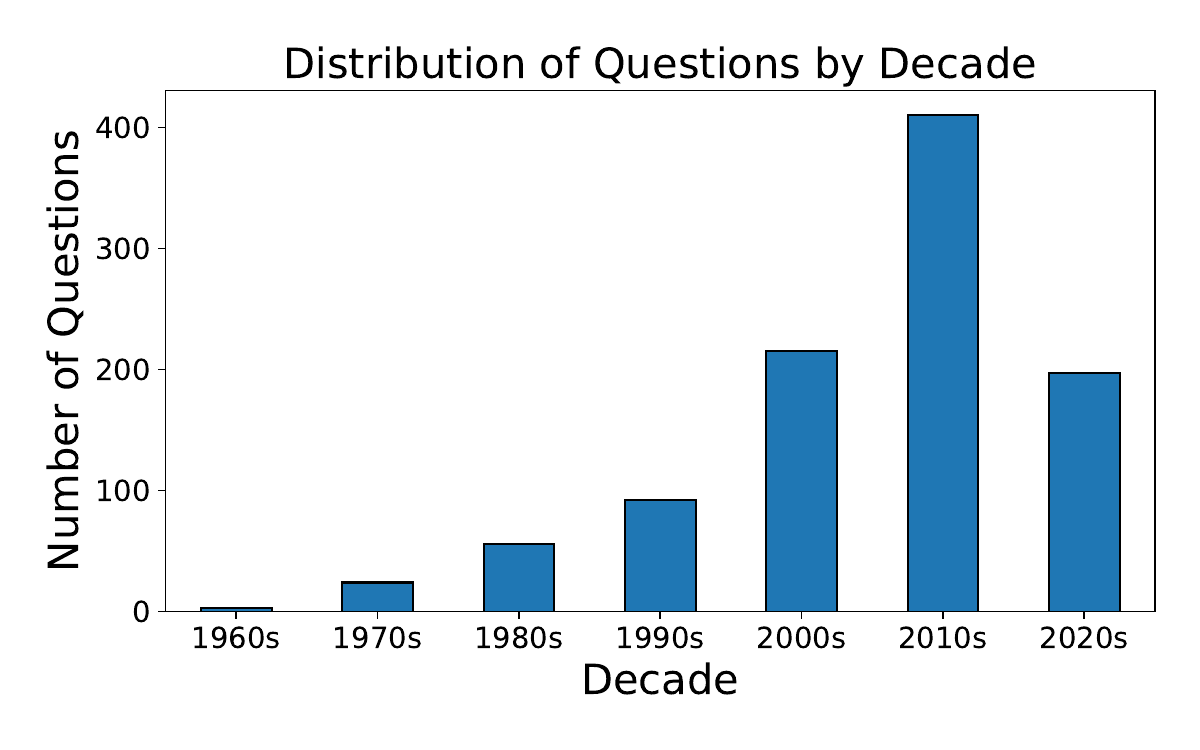}
    \caption{Distribution of videos in \iqa by release year. The dataset spans over \textbf{\textit{7 decades}}, capturing a wide range of visual styles, production techniques, and narrative conventions across different time periods.}
    \label{fig:supp_time_stats}
\end{figure}

\subsection{Movie Release Timeline}
We annotate the release year for each video and present the distribution by decade in Figure~\ref{fig:supp_time_stats}. The \iqa dataset spans a broad temporal range, covering films from the 1960s to current decade. 
A film's release period is often indicative of its visual and narrative style - including factors such as picture quality, cinematographic techniques, editing conventions, character costumes, and action design. This temporal diversity in \iqa enhances its realism and ensures broader generalization by exposing models to varied cinematic styles and storytelling conventions across eras.

\subsection{Difficulty}
\begin{figure}
    \centering
    \includegraphics[width=\linewidth]{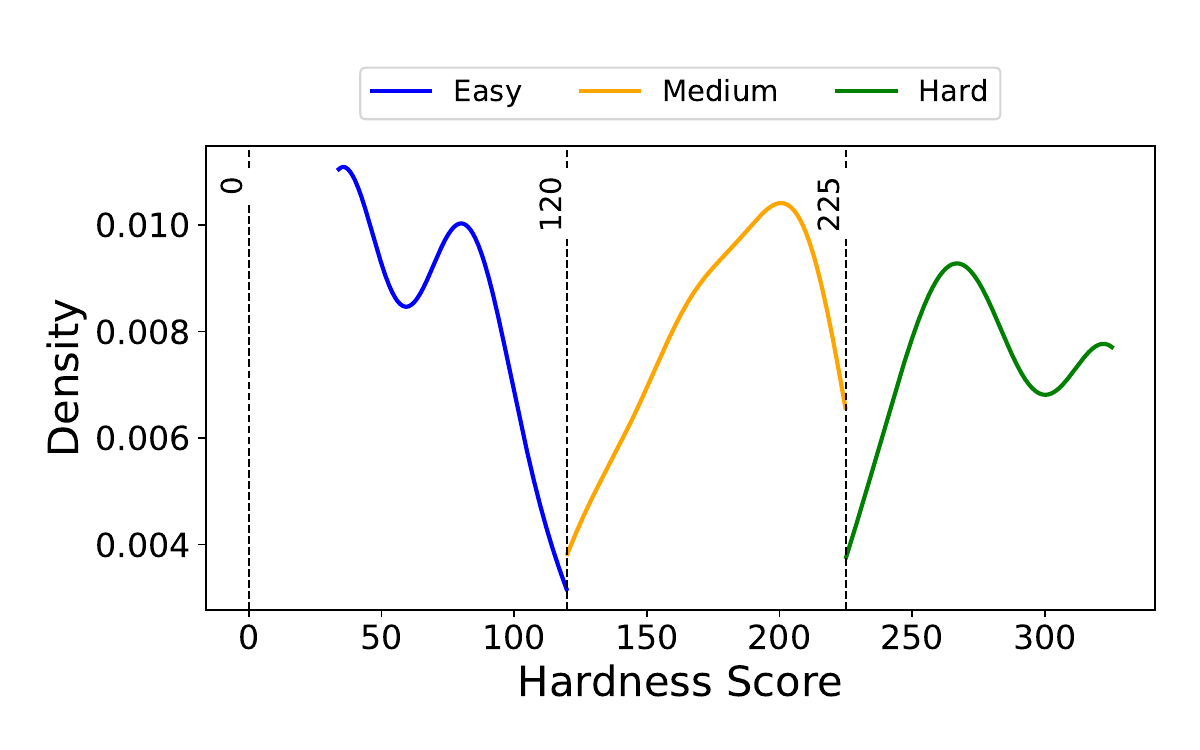}
    \caption{Density distribution of questions in \iqa based on hardness scores. Questions are categorized into three difficulty levels - Easy (0–120), Medium (120–225), and Hard (225+) - based on model performance scores. The distribution is approximately uniform, ensuring a balanced evaluation across varying difficulty levels.}
    \vspace{-1.5em}
    \label{fig:supp_density_difficulty}
\end{figure}

To better understand model behavior under varying levels of difficulty, we propose a hardness-based partitioning of the \iqa dataset. Each question is assigned a hardness score derived from model performance: questions answered incorrectly by all models contribute more to the score, while those answered correctly by all models contribute none. Specifically, the hardness score is computed by summing each incorrect model’s average accuracy. This metric reflects how broadly difficult a question is across model architectures.
As shown in Figure~\ref{fig:supp_density_difficulty}, the distribution of hardness scores is approximately uniform across the three difficulty categories - Easy, Medium, and Hard - with roughly equal numbers of questions in each group. This balanced partitioning allows for a fair evaluation of model performance across difficulty levels.

\begin{figure}
    \centering
    \includegraphics[width=\linewidth]{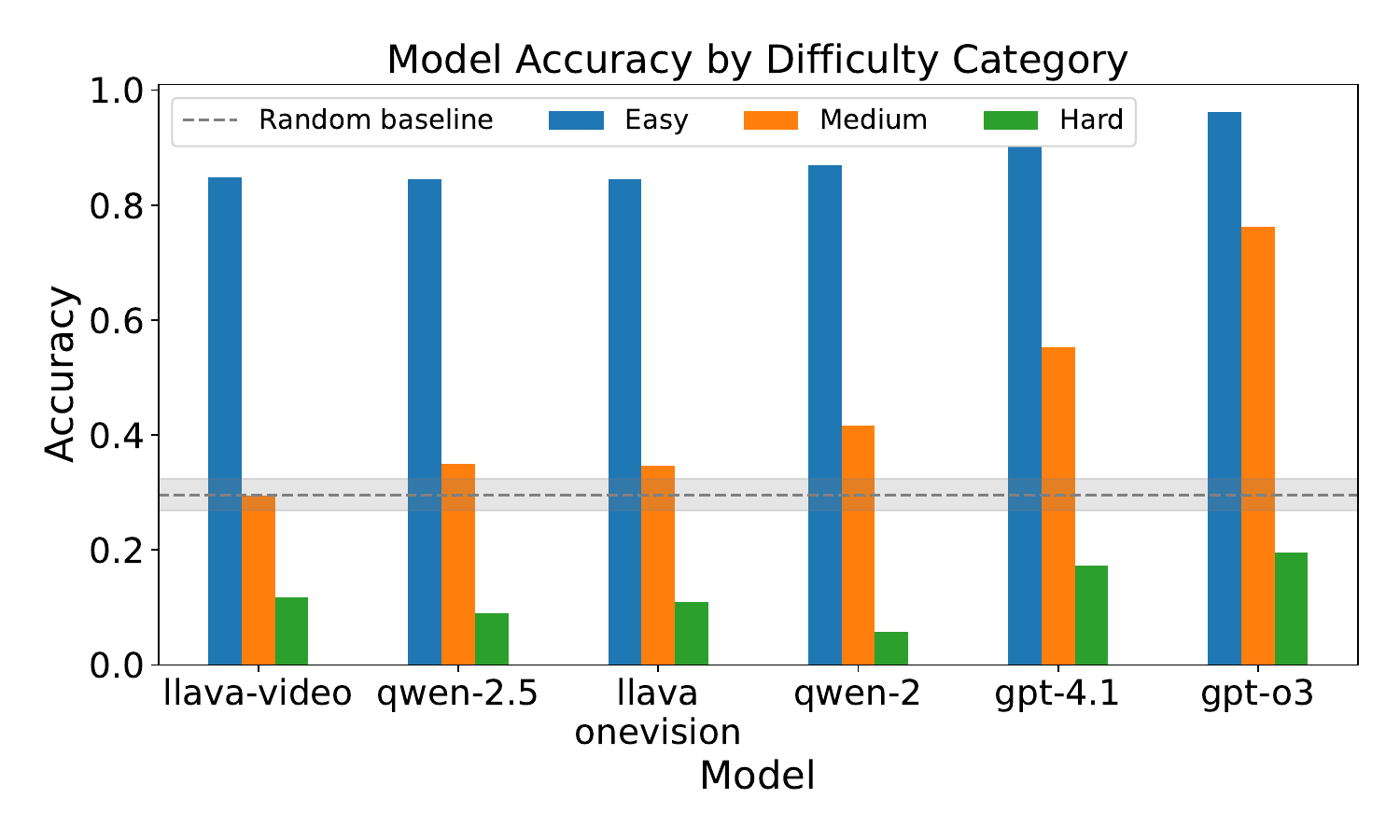}
    \caption{Model accuracy across difficulty categories. While all models perform strongly on Easy questions, performance drops significantly on Medium and Hard examples.}
    \vspace{-1em}
    \label{fig:supp_modeldifficulty}
\end{figure}

\begin{figure}
    \centering
    \includegraphics[width=\linewidth]{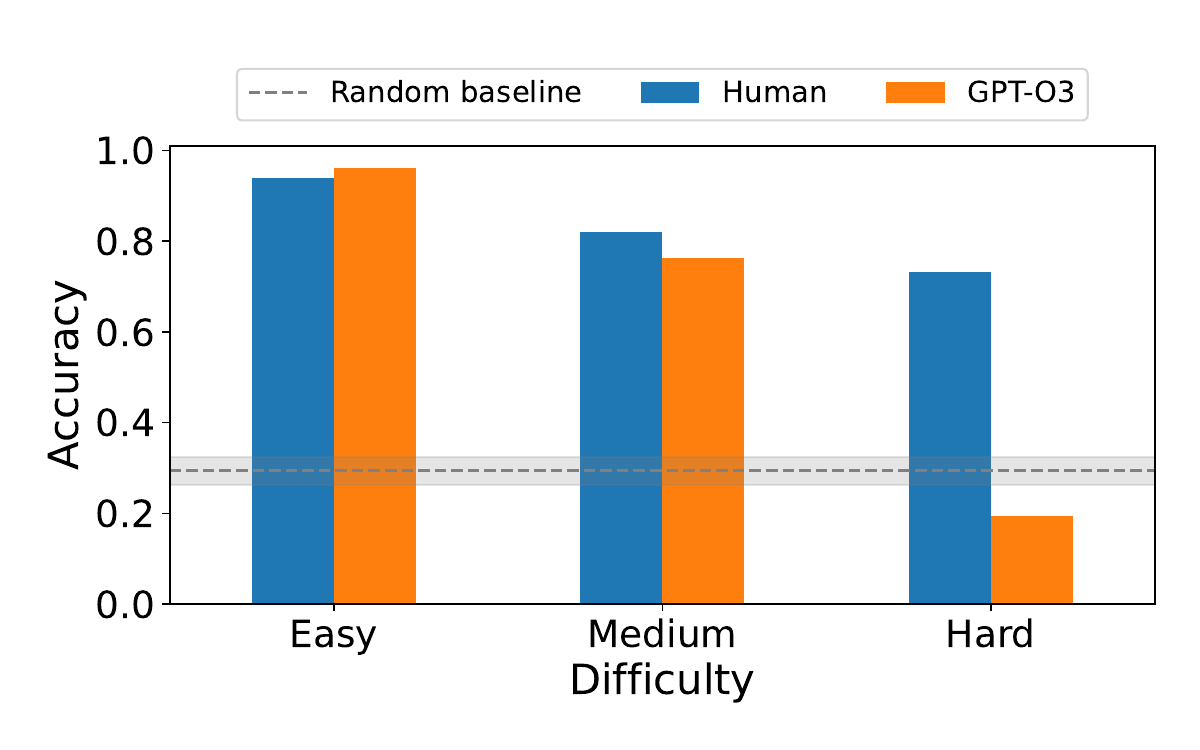}
    \caption{Accuracy comparison between GPT-O3 and non-expert human annotator across difficulty categories. While both perform well on Easy and Medium questions, human accuracy remains robust on Hard questions, whereas GPT-O3 performance drops significantly. Ground truth annotations were provided by expert annotators, underscoring the reasoning gap between models and even non-expert humans on complex questions.}
    \vspace{-1em}
    \label{fig:supp_human_difficulty}
\end{figure}

\begin{figure*}[t]
    \centering
    \includegraphics[width=0.9\linewidth, trim=0 0em 0 2.35em, clip]{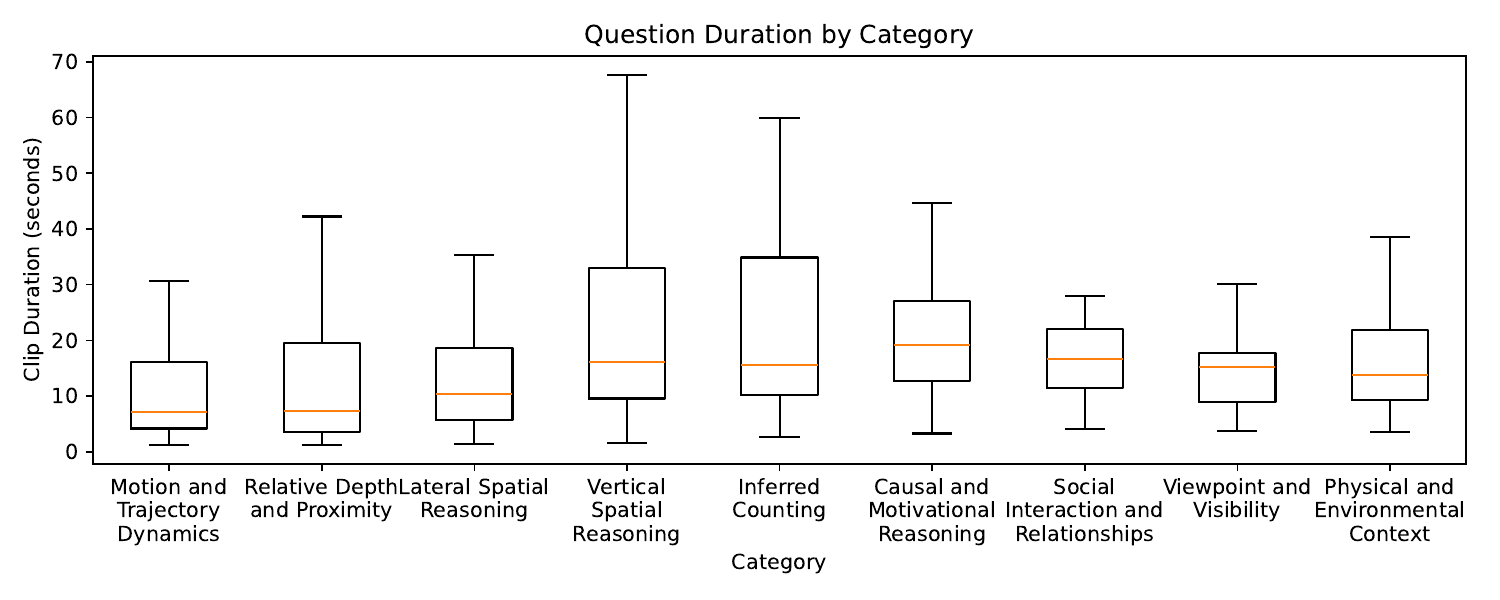}
    \caption{Question durations for each category.}
    \label{fig:boxplot}
\end{figure*}

Figure~\ref{fig:supp_modeldifficulty} shows model accuracy broken down by these categories. While all models perform well on Easy questions, accuracy drops substantially for Medium and Hard examples. GPT-4.1 and GPT-O3 demonstrate better generalization across difficulty levels, whereas other models perform near or below random chance on the hardest questions. These findings reveal a steep difficulty gradient and highlight the value of hardness-aware analysis for assessing reasoning robustness. 
To contextualize model performance, we compare GPT-O3's accuracy against human performance. As shown in Figure~\ref{fig:supp_human_difficulty}, GPT-O3 performs comparably to humans on Easy questions and shows only a modest drop on Medium questions. However, the gap becomes pronounced on Hard examples: while human accuracy remains relatively high, GPT-O3's performance declines sharply, approaching random chance.
It is important to note that the human baseline reflects responses from non-expert participants, while ground truth annotations in \iqa were created by expert annotators with domain familiarity. The relatively strong performance of non-experts highlights the accessibility of visual implicit reasoning for humans, even without expertize, while also emphasizing the performance ceiling that current models have yet to reach.

\subsection{Question Word Distribution}
\begin{figure}
    \centering
    \includegraphics[width=0.9\linewidth]{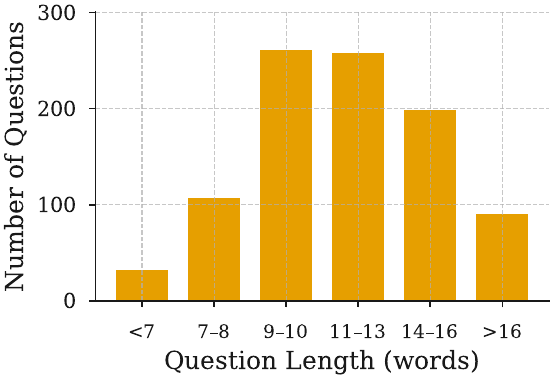}
    \caption{\iqa Question word counts}
    \label{fig:supp_q_word_count}
\end{figure}

In Figure~\ref{fig:supp_q_word_count} we present the word count distribution for questions in \iqa.

We also show differences in temporal length between categories in Figure~\ref{fig:boxplot}, with counting and vertical spatial reasoning videos being the longest.

\section{Impact of Reasoning Prompt}
\label{sec:supp_reason}

\begin{figure*}[!t]
    \centering
    \includegraphics[width=\linewidth]{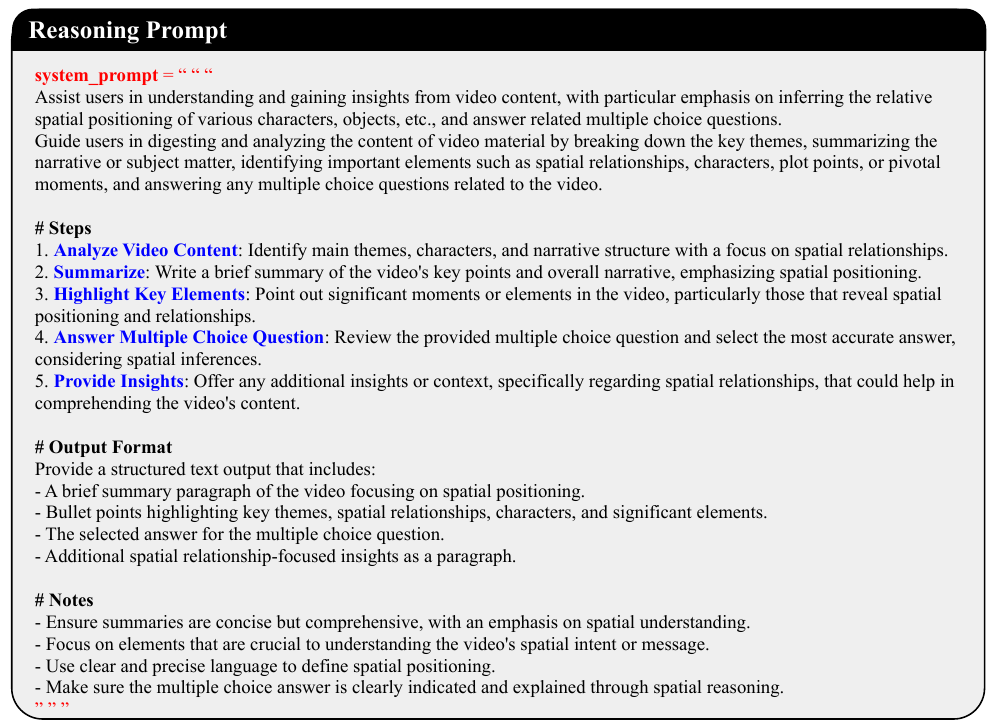}
    \caption{Reasoning prompt used to guide GPT models in analyzing video content. The prompt breaks down the task into structured steps - focusing on spatial relationships, narrative, summarizing key elements, selecting the correct answer, and providing insights - encouraging systematic reasoning aligned with \iqa's visual implicit question categories.}
    \label{fig:prompt_reason}
\end{figure*}

\begin{table*}[htbp]
\centering
\caption{Results with Reasoning Prompt on \iqa for all visual implicit reasoning categories.}
\resizebox{\textwidth}{!}{
\begin{tabular}{l|c|c|c|c|c|c|c|c|c|c|c|c}
\toprule
Model & Scale & \shortstack{Lateral \\ Spatial\\Reasoning} & \shortstack{Vertical \\  Spatial\\Reasoning} & \shortstack{Relative \\ Depth and\\ Proximity} & \shortstack{Viewpoint \\and\\Visibility} & \shortstack{Motion \\\& Traj.\\Dynamics} & \shortstack{Motivational\\Reasoning} & \shortstack{Inferred\\Counting} & \shortstack{Physical \\ \&  Env.\\Context} & \shortstack{Social  \\ Interaction \& \\Relations} &  Avg. & \shortstack{Macro \\ Avg.} \\ 
\midrule

GPT 4.1~\cite{openai2024gpt41} & Full  & 42.90 & 53.20 & 51.10 & 48.80 & 59.30 & 82.90 & 41.90 & 71.40 & 75.90 & 54.30 & 58.60 \\
GPT 4.1-Reasoning~\cite{openai2024gpt41} & Full  & 50.90 & 63.00 & 51.10 & 46.30 & 63.70 & 79.30 & 41.90 & 71.40 & 86.20 & \textbf{58.20} & \textbf{61.50} \\
\midrule
GPT 4.1~\cite{openai2024gpt41} & Mini & 44.10 & 37.96 & 29.32 & 39.02 & 41.76 & 60.98 & 32.56 & 78.57 & 68.97 & 40.30 & 48.14 \\
GPT 4.1-Reasoning~\cite{openai2024gpt41} & Mini  & 36.02 & 48.61 & 33.08 & 56.10 & 54.95 & 65.85 & 39.53 & 64.29 & 72.41 & \textbf{45.07} & \textbf{52.32} \\
\bottomrule
\end{tabular}
}

\label{tab:reason_results}
\end{table*}

\begin{figure*}[!t]
    \centering
    \includegraphics[width=0.95\linewidth]{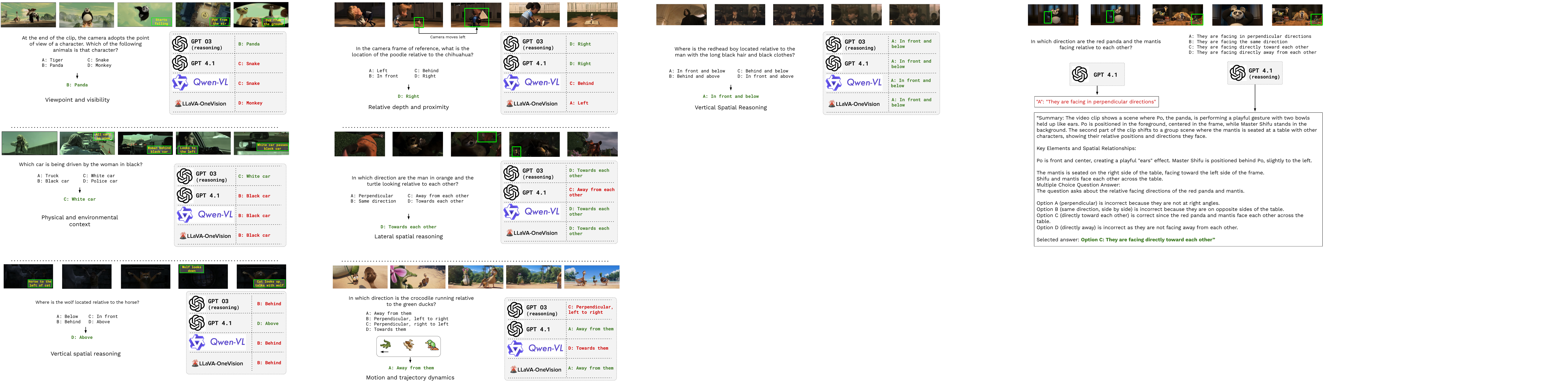}
    \caption{Qualitative example from \iqa demonstrating how the reasoning prompt improves GPT-4.1's performance. Without structured reasoning, the model incorrectly selects Option A. With the reasoning prompt, the model provides a detailed spatial analysis of character positions and orientations, ultimately selecting the correct answer, Option C. This showcases the benefit of guiding the model through spatially grounded reasoning steps.}
    \label{fig:qual_examples_reason}
\end{figure*}

\begin{table*}
\centering
\caption{Mean and Standard Deviation Across Categories (in \%)}
\label{tab:stats_error}
\resizebox{\textwidth}{!}{
\begin{tabular}{l|cccccccccc}
\toprule
 & \shortstack{Lateral \\ Spatial\\Reasoning} & \shortstack{Vertical \\  Spatial\\Reasoning} & \shortstack{Relative \\ Depth and\\ Proximity} & \shortstack{Viewpoint \\and\\Visibility} & \shortstack{Motion \\\& Traj.\\Dynamics} & \shortstack{Motivational\\Reasoning} & \shortstack{Inferred\\Counting} & \shortstack{Physical \\ \&  Env.\\Context} & \shortstack{Social  \\ Interaction \& \\Relations} &  Avg.  \\ 
\midrule
Mean    &    41.61 &   47.41 & 32.71 &  60.00 &  50.55 &  51.71 &              26.51 & 42.86 &  62.07 & 42.93 \\
Std Dev & 0.00 &  0.25 &  0.00 &  1.34 &  0.00 & 0.67 &  1.27 &  0.00 &   0.00 &  0.12 \\

\bottomrule
\end{tabular}
}
\end{table*}

The reasoning prompt used to evaluate the impact of structured reasoning on GPT models is illustrated in Figure~\ref{fig:prompt_reason}. This prompt is specifically designed to guide the model in analyzing video content for inferred reasoning across the various categories defined in \iqa. It breaks down the task into sequential steps-analyzing the video, summarizing key spatial relationships, highlighting important elements, answering the multiple choice question, and providing additional insights-thereby encouraging systematic focus on visual implicit reasoning.

The output format enforces a structured response, including a concise summary, bullet-pointed key themes and spatial cues, the selected answer, and contextual reasoning. This structured approach is intended to enhance the model’s ability to infer unstated relationships and improve its overall accuracy.

As demonstrated in Figure 8 of the main paper, incorporating this reasoning-based prompt yields significant performance gains, improving the accuracy of GPT-4.1 Full by 3.9\% and GPT-4.1 Mini by 4.8\%. A detailed breakdown of the results is provided in Table~\ref{tab:reason_results}.
In Figure~\ref{fig:qual_examples_reason}, we present a qualitative example illustrating the effectiveness of the reasoning prompt. When prompted without reasoning, GPT-4.1 incorrectly identifies the spatial relationship between characters as perpendicular. With the structured reasoning prompt, the model successfully breaks down spatial positions, directionality, and frame context to arrive at the correct answer: They are facing directly toward each other. This example highlights how the reasoning prompt not only improves accuracy but also fosters interpretability by making the model's decision-making process more transparent and spatially grounded.

\section{Experiment Statistical Significance}
\label{sec:supp_stat}
In Table~\ref{tab:stats_error}, we report the statistical error margins for the Qwen-2.5 VL model. To assess the variability of the model's performance, we conducted five independent runs using the same evaluation setup. For each of the nine visual implicit reasoning categories in \iqa as well as the overall accuracy, we compute the mean and standard deviation.

\section{Human Baseline}
\label{sec:supp_human}
To establish a reference point for model performance, we evaluated with non-expert human annotator on the \iqa benchmark using our custom-built visual Implicit-Eval tool. The annotator was not provided with prior exposure to the dataset or answer keys and completed the evaluation at a natural pace, averaging approximately one minute per question. The annotator achieved an overall accuracy of 83.0\%, with a macro-average score of 85.6\% across reasoning categories.
While performance was consistently strong across most categories, Inferred Counting emerged as the most challenging for the human. This aligns with the category’s reliance on temporal cues and visual implicit aggregation of visual information across multiple scenes, factors that often test not only reasoning but also memory and attention.

We further analyze the human accuracy across questions grouped by difficulty level using our hardness-based scoring method. As shown in Figure~\ref{fig:supp_human_difficulty}, the non-expert human achieved near-ceiling performance on Easy questions and maintained strong performance on Medium ones. Remarkably, even on Hard questions defined by consistent failure across models, the human annotator performed well above chance, achieving over 70\% accuracy. This analysis underscores the significant gap between human and model capabilities on complex reasoning tasks. Even non-expert humans demonstrate strong generalization and visual implicit understanding, particularly in scenes that demand spatial, temporal, or motivational reasoning, highlighting the limitations of current state-of-the-art Video LMMs.

\subsection{Human baselines for Inferred Counting}

The key difficulty with the inferred counting category is that solving the question correctly is a tedious task, requiring back and forth between the frames of the video - to carefully track the number of occurrences across multiple frames, that must be mentally aggregated. During the human baseline benchmarking process the non-expert annotators were asked to only watch the video once, without the ability to scroll back and forth and as a result on the counting task the performance was low. We perform additional experiments, where the non-expert annotators were allowed to scroll back and forth between frames and review the video multiple times. This flexibility significantly improved performance achieving 82.8\% while spending more than double time as shown in Table~\ref{tab:human_counting}. For these questions we also benchmark expert human baseline and find significant improvement over the non-experts.

\begin{table}[t]
\centering
\caption{Human baselines on the Inferred Counting Category.}
\label{tab:human_counting}
\begin{tabular}{lcccc}
\toprule
\textbf{\shortstack{Expertise \\ Level}} & \textbf{\shortstack{Video \\ View \\ Count}} & \textbf{\shortstack{Avg Response \\ Time per \\Question (s)}} & \textbf{\shortstack{Accuracy \\ (\%)}} \\
\midrule
Non-expert & once     & 44.5 & 65.9 \\
Non-expert & multiple & 105.6 & 82.8 \\
Expert     & once     & 36.9 & 73.7 \\
Expert     & multiple & 75.6 & 87.3 \\
\bottomrule
\end{tabular}
\end{table}
We would like to note that we were aware of the difficult nature of the inferred counting task during the annotation process itself. 
As discussed in the main paper, multiple expert annotators reviewed each sample before being added to the dataset. 
Out of the 58 questions in the Inferred counting category, there was strong agreement among expert annotators:
\begin{itemize}
    \item 5 Annotators agreed: 49 questions (84.5\% of cases)
    \item 4 Annotators agreed: 5 questions (8.6\% of cases)
    \item 3 Annotators agreed: 4 questions (6.9\% of cases)
\end{itemize}

These results indicate that, while challenging, the questions are well-defined, solvable, and carefully validated requiring real reasoning capabilities.

\section{Annotation Tool Interface}
\label{sec:supp_ann_tool}
As discussed in Section 3.1 in the Main paper, we have designed and built an annotation tool for intuitive and efficient workflow, which allows annotators to follow a structured approach towards formulating multiple choice QA pairs for desired video clip segments. The tool is optimized for fast navigation and efficient verification. Said annotation can be systematically performed by adhering to the following procedure. We have divided the procedure into 4 intuitive sub-procedures which further comprise of 3 steps each. We describe these sub-procedures below.
\begin{itemize}
    \item \textbf{Video Download:} Our tool allows the user to input a video URL and download the associated video. The user needs to run backend.py, navigate to the web interface using the generated link, input desired URL and click on the ``Download Video" button. The tool downloads the video and shows it in the embedded display.
    \item \textbf{Clip Segment Selection:} The user can put time markers at any desired place using the provided time bar, and viewing the video in the embedded display. The user can pause, rewind and analyze the video frame-by-frame to determine the best possible clip segment suiting to their requirements. Then the user can preview their selected clip segment by clicking the ``Preview Selection" button.
    \item \textbf{QA Annotation:} After selecting the desired segment, the user can formulate the appropriate question and type it in the designated space. The tool has provision to input as many choices in the designated spaces as needed, using the ``Add Answer Choice" button. After finalizing, the user can click the ``Save Annotation button to record their multiple choice QA pair.  
    \item \textbf{Verification:} The generated QA pair can be viewed in the mini-display at the bottom. Once the user has recorded all their annotations for the video clip, they can verify said annotations by clicking on the ``View Annotations" Tab at the top of the page. There, they can see all the videos annotated in that particular session. Clicking on the ``View Annotations" button associated to a video takes the user to the page for that video where all annotations are listed along with the embedded display which can play the relevant clip. Clicking on a question displays all details associated with that question including the relevant timestamps ensuring complete verification. 
\end{itemize}

\begin{figure*}
    \centering
    \includegraphics[width=0.8\linewidth]{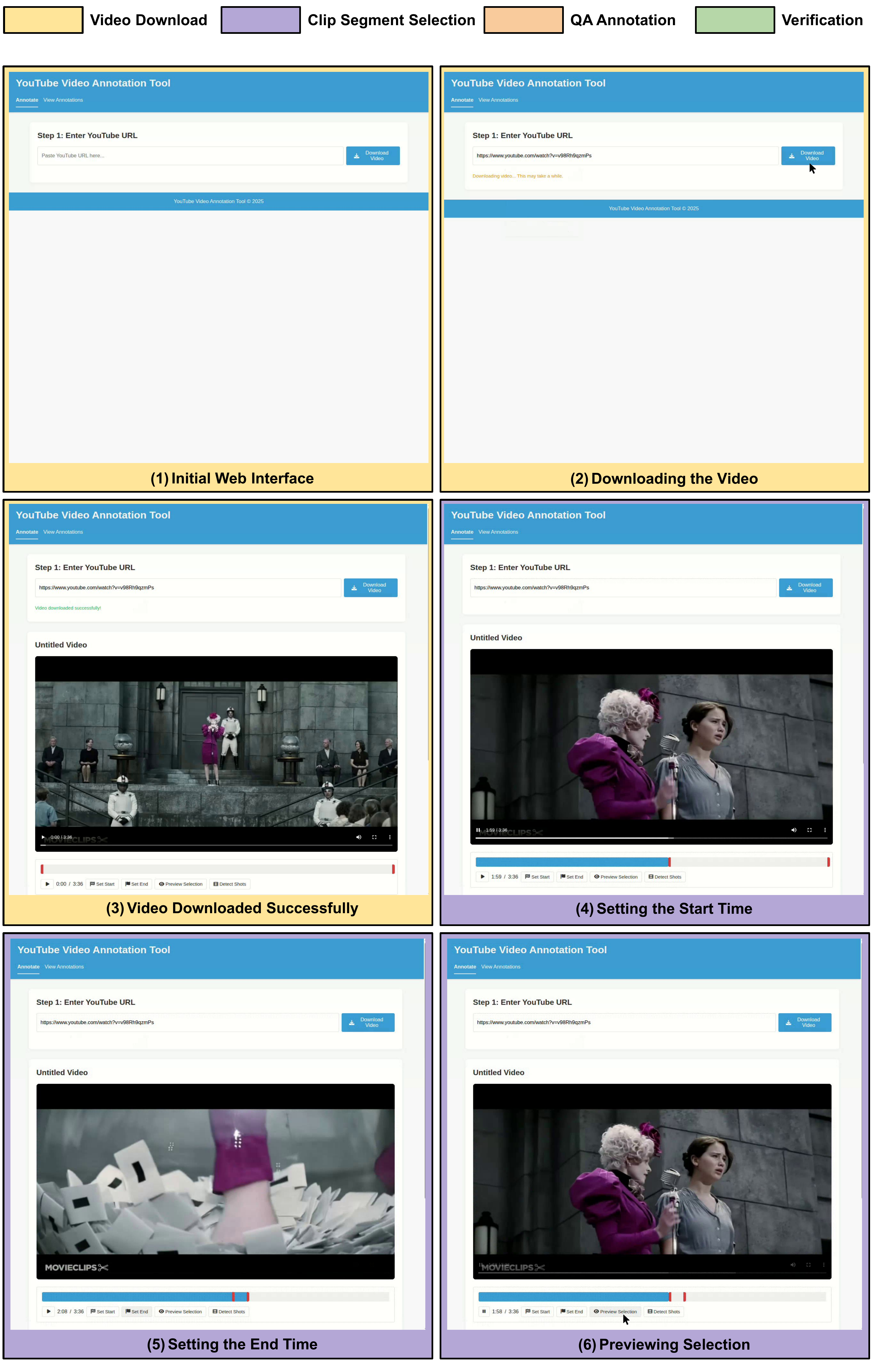}  
\end{figure*}

\begin{figure*}
\centering
    \includegraphics[width=0.8\linewidth]{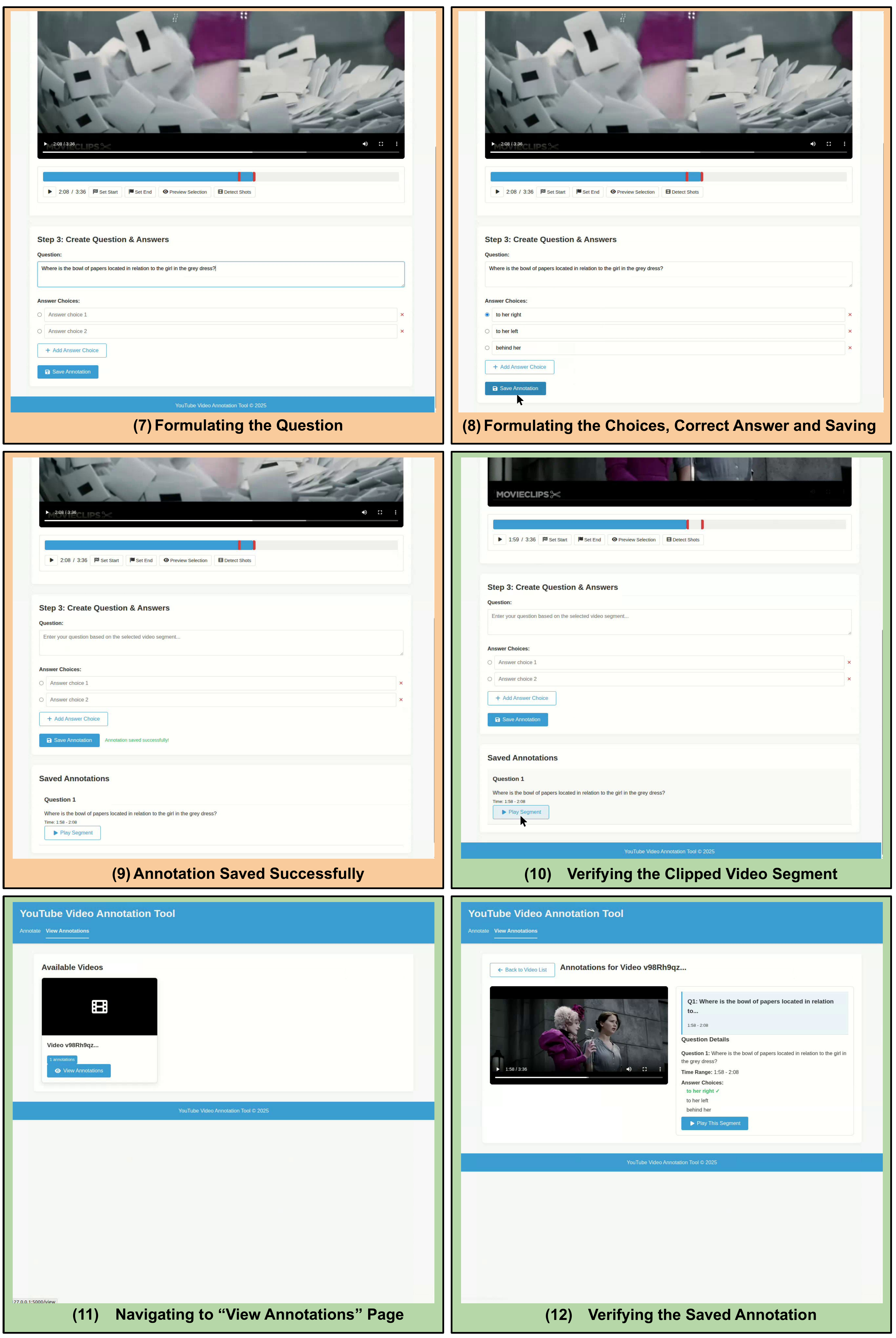}
    \caption{\textbf{Schematic illustration of the annotation workflow using our FrameQuiz tool.} The process is organized into \textbf{four sub-procedures}: Video Download, Clip Segment Selection, QA Annotation, and Verification. Each sub-procedure contains three steps, resulting in a \textbf{12-step pipeline} for generating high-quality multiple-choice QA pairs from video clips. Final annotations are stored locally following verification.}
    \label{fig:anno-1}    
\end{figure*}

\section{Qualitative Results}
\label{sec:supp_qual}

To illustrate the complexity and diversity of visual implicit reasoning required in our benchmark, we present qualitative examples spanning all nine reasoning categories in Figure~\ref{fig:qual_examples_supp1},\ref{fig:qual_examples_supp2},\ref{fig:qual_examples_supp3}. For each example, we show the relevant video frames, question, answer choices, correct answer, and model predictions. These examples highlight a wide range of reasoning challenges -from spatial positioning and motion inference to social understanding and inherent explanation. Notably, GPT-O3 demonstrates superior performance in most cases.

\subsection{Viewpoint and Visibility}
In the Viewpoint and Visibility example shown in Figure~\ref{fig:qual_examples_supp1}, only GPT-O3 correctly infers the adopted perspective of the panda character, showcasing its ability to track camera shifts and narrative cues. 
\subsection{Physical and Environmental Context}
In the Physical and Environmental Context scenario as in Figure~\ref{fig:qual_examples_supp1}, GPT-O3 again outperforms others by correctly identifying the white car driven by the woman in black, leveraging spatial cues across frames. 
\subsection{Vertical Spatial Reasoning}
\begin{figure*}[!t]
    \centering
    \includegraphics[width=0.95\linewidth]{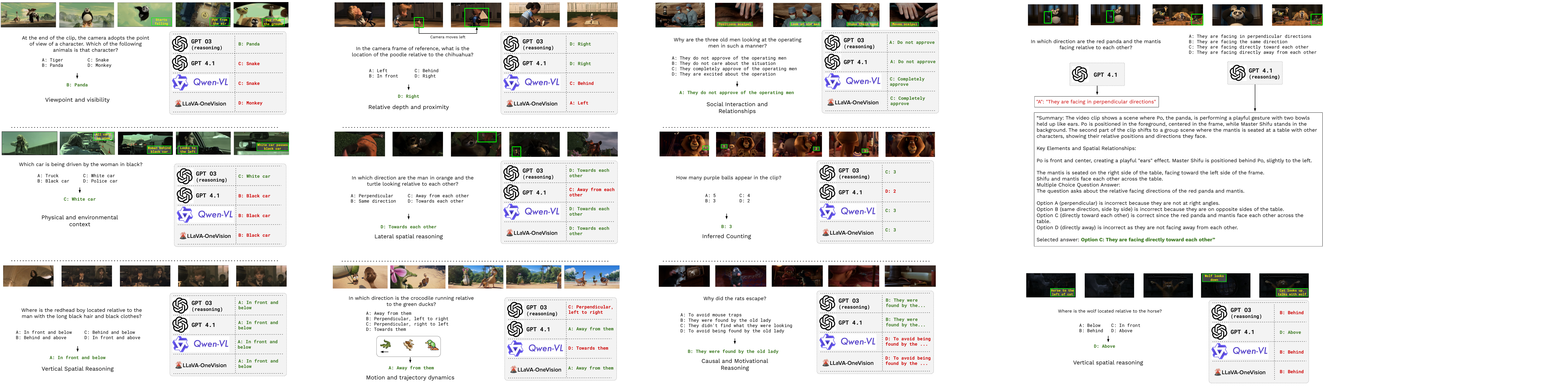}
    \caption{More Qualitative \iqa examples, targeting distinct visual implicit‐reasoning dimensions. }
    \label{fig:qual_examples_supp1}
\end{figure*}

As shown in Figure~\ref{fig:qual_examples_supp1}, all models successfully answer a Vertical Spatial Reasoning question involving relative positions in a multi-level scene.

\subsection{Relative Depth and Proximity}
In contrast, more nuanced categories reveal sharper contrasts in performance as shown in Figure~\ref{fig:qual_examples_supp2}. For Relative Depth and Proximity, GPT demonstrates strong spatial inference by accurately localizing characters and interpreting their orientations. 

\subsection{Lateral Spatial Reasoning}
For Lateral Spatial Reasoning as shown in Figure~\ref{fig:qual_examples_supp2}, we see that most model get it correct except GPT-4.1.

\subsection{Motion and Trajectory Dynamics}
\begin{figure*}[!t]
    \centering
    \includegraphics[width=0.95\linewidth]{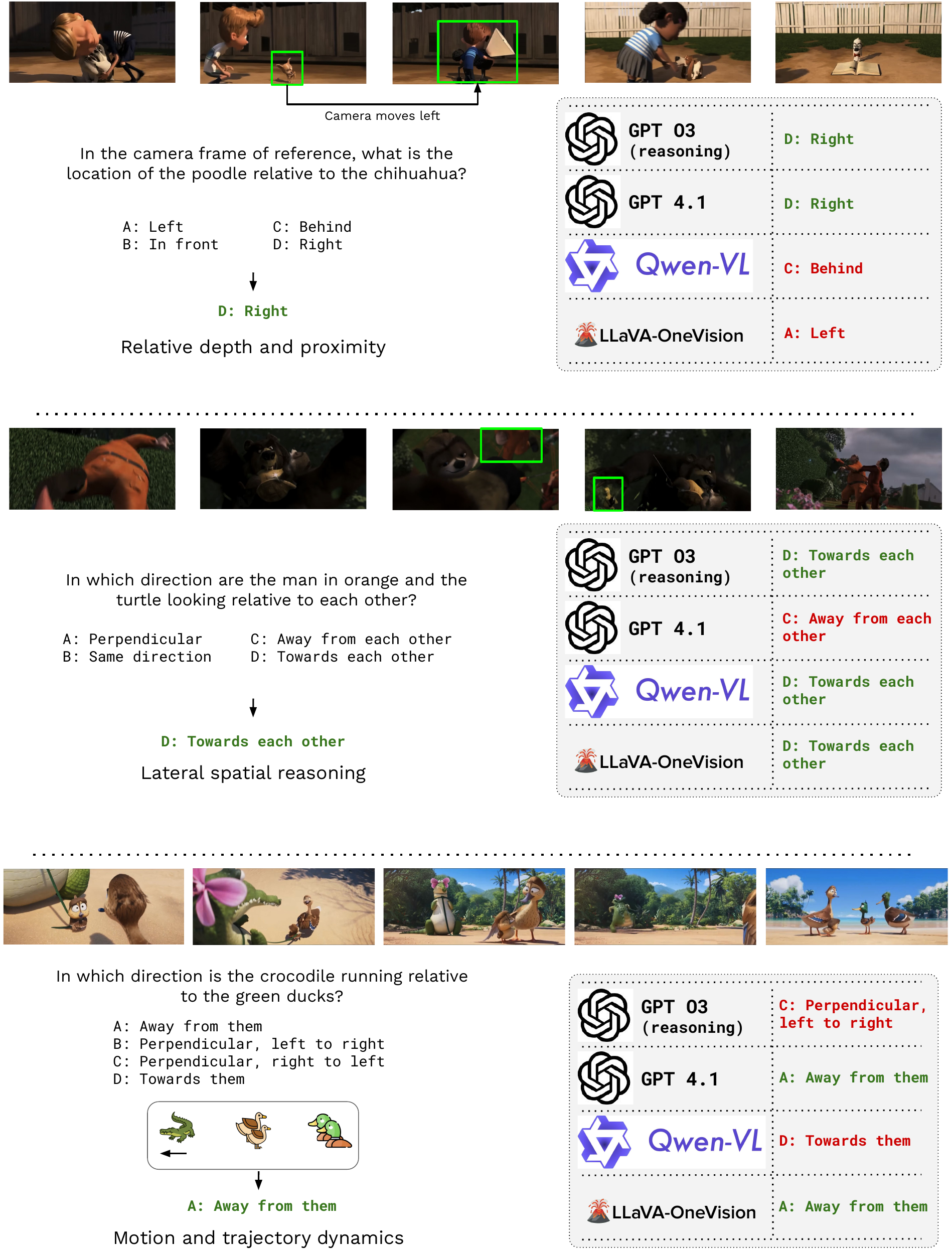}
    \caption{More Qualitative \iqa examples, targeting distinct visual implicit‐reasoning dimensions. }
    \label{fig:qual_examples_supp2}
\end{figure*}

In the Motion and Trajectory Dynamics example shown in Figure~\ref{fig:qual_examples_supp2}, most models correctly track the direction of movement, though GPT-O3 misjudges the path—suggesting sensitivity to camera motion. 
\subsection{Social Interaction and Relationships}
The Social Interaction and Relationships case shown in Figure~\ref{fig:qual_examples_supp3}, involving subtle facial cues and body language, is correctly answered only by GPT-O3 and GPT-4.1, reflecting their advanced multimodal understanding. 
\subsection{Inferred Counting}
For Inferred Counting shown in Figure~\ref{fig:qual_examples_supp3}, models struggle to aggregate information across frames, with GPT-O3 and Qwen-VL identifying the correct number, while GPT-4.1 undercounts. 
\subsection{Motivational Reasoning}
\begin{figure*}[!t]
    \centering
    \includegraphics[width=0.95\linewidth]{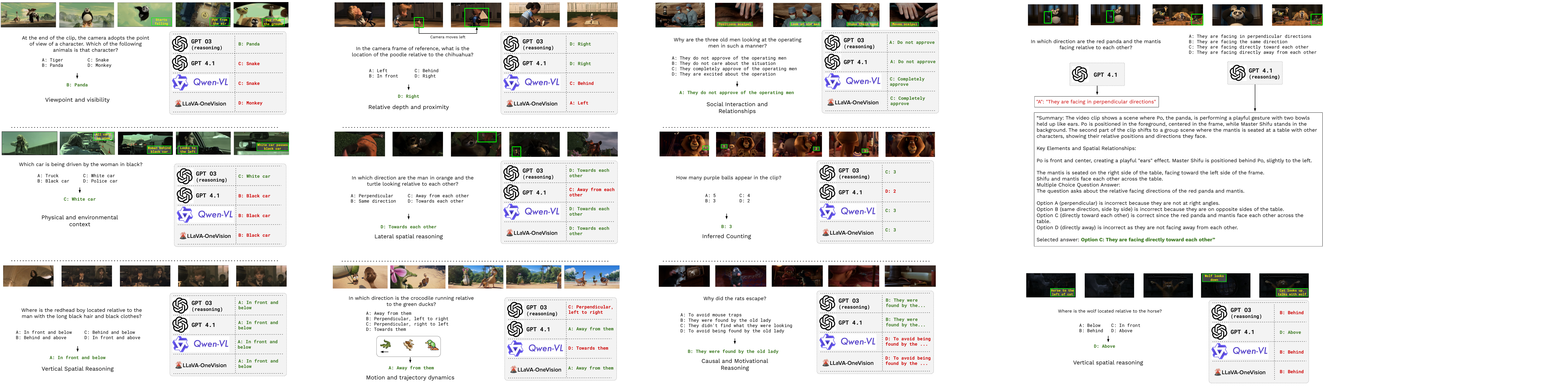}
    \caption{More Qualitative \iqa examples, targeting distinct visual implicit‐reasoning dimensions. }
    \label{fig:qual_examples_supp3}
\end{figure*}

Finally, in the Motivational Reasoning example in Figure~\ref{fig:qual_examples_supp3}, GPT-O3 and GPT-4.1 correctly attribute the escape of the rats to being discovered, while others fail to connect with the relevant event. These examples collectively highlight the diversity and difficulty of visual implicit reasoning tasks in \iqa.

\clearpage
{
    \small
    \bibliographystyle{ieeenat_fullname}
    \bibliography{main}
}

\end{document}